\pdfoutput=1
\documentclass[a4paper,fleqn]{cas-sc}
\usepackage[authoryear]{natbib}
\usepackage{amsmath,amssymb,booktabs,tabularx,array,multirow,makecell}
\usepackage{algorithm,algpseudocode}
\usepackage{graphicx,xcolor,colortbl,placeins,enumitem,microtype}
\usepackage{threeparttable,pifont}
\newcommand{\cmark}{\ding{51}}
\newcommand{\xmark}{\ding{55}}
\newcommand{\second}[1]{\underline{#1}}
\newcommand{\ci}[1]{{\color{black!55}[#1]}}
\usepackage{caption}

\makeatletter
\RenewDocumentEnvironment{figure}{O{tbp}}{\@float{figure}[#1]}{\end@float}
\RenewDocumentEnvironment{figure*}{O{tbp}}{\@dblfloat{figure}[#1]}{\end@dblfloat}
\makeatother

\usepackage{tikz}
\usetikzlibrary{arrows.meta,positioning,calc,fit,backgrounds}
\newcolumntype{Y}{>{\raggedright\arraybackslash}X}
\newcolumntype{P}[1]{>{\raggedright\arraybackslash}p{#1}}
\newlist{lightitemize}{itemize}{1}
\setlist[lightitemize]{label=\textbullet,leftmargin=1.2em,itemsep=0.25em,topsep=0.3em}
\definecolor{sageblue}{HTML}{28465A}
\definecolor{sagepale}{HTML}{EFF4F6}
\definecolor{sagesand}{HTML}{F6F1E8}
\hypersetup{colorlinks=true,citecolor=black,linkcolor=black,urlcolor=sageblue}
\newcommand{\sagepdfinfo}{\hypersetup{pdfauthor={Jiaxi Liu, Keshu Wu, Weizhe Tang, Bofeng Cao, Sikai Chen, Xiaopeng Li, Steven T. Parker, Bin Ran},pdfsubject={Work-zone pathway prediction under temporary traffic control},pdfkeywords={Roadway work zones, Temporary traffic control, Automated vehicles, Pathway prediction, Vision-language models, Traffic control standards}}}
\sagepdfinfo
\renewcommand{\topfraction}{0.93}
\renewcommand{\bottomfraction}{0.75}
\renewcommand{\textfraction}{0.06}
\renewcommand{\floatpagefraction}{0.75}
\begin{document}
\let\WriteBookmarks\relax
\shorttitle{SAGE for work-zone pathway prediction}
\shortauthors{Liu et al.}
\title[mode=title]{SAGE: A hybrid vision--language and geometry framework for work-zone pathway prediction under temporary traffic control}
\author[1]{Jiaxi Liu}
\author[1]{Keshu Wu}
\cormark[1]
\ead{kwu84@wisc.edu}
\author[1]{Weizhe Tang}
\author[1]{Bofeng Cao}
\author[1]{Sikai Chen}
\author[1]{Xiaopeng Li}
\author[1]{Steven~T. Parker}
\author[1]{Bin Ran}
\affiliation[1]{organization={Department of Civil \& Environmental Engineering, University of Wisconsin--Madison},city={Madison},postcode={53706},state={Wisconsin},country={United States}}
\cortext[cor1]{Corresponding author}
\begin{abstract}
Temporary traffic control redirects vehicles through work zones, requiring automated vehicles to connect an interpretation of the visible devices to a continuous passing path. Geometric predictors provide several plausible paths, yet selecting the one consistent with the temporary control remains difficult. We propose SAGE (Semantic-to-Action Grounded Execution), a read--decide--execute--check framework that separates path construction, path preference, and control compliance within a shared candidate space. The central idea is to use a vision--language model's (VLM's) trajectory as a preference over intact geometric candidates. A role-decoupled generator constructs the candidates, and rank fusion combines each candidate's geometric score with its agreement with the VLM trajectory. Supervision structured around the Manual on Uniform Traffic Control Devices (MUTCD) trains SAGE-VLM to state the lane status and hazard side before its passage decision and trajectory. The predicted hazard side enters image-space predicates derived from MUTCD provisions, which support a final intervention bounded in score and displacement. On 2{,}306 real-world ROADWork validation episodes, the candidates include a path within 10\,px of the human demonstration in 87\% of episodes, yet selection regret accounts for 53\% of the geometric ranker's error. SAGE reduces average and final displacement errors by 8.3\% and 12.0\% relative to the ranker and hard-constraint violations by 37\%. The trajectory opinion improves all four tested generator--selector combinations. SAGE is comparable to state-of-the-art VLM modules in accuracy, with a faster-responding VLM, and outperforms four other adapted driving VLMs in displacement error and safety measures. Project page: \url{https://hhjj233.github.io/SAGE-page/}.
\end{abstract}
\maketitle
\sagepdfinfo
\renewcommand{\topfraction}{0.94}
\renewcommand{\bottomfraction}{0.8}
\renewcommand{\textfraction}{0.04}
\renewcommand{\floatpagefraction}{0.7}
\setcounter{topnumber}{4}
\setcounter{totalnumber}{6}
\section{Introduction}
\label{sec:intro}

\begin{figure}[t]
\centering
\includegraphics[width=0.78\linewidth]{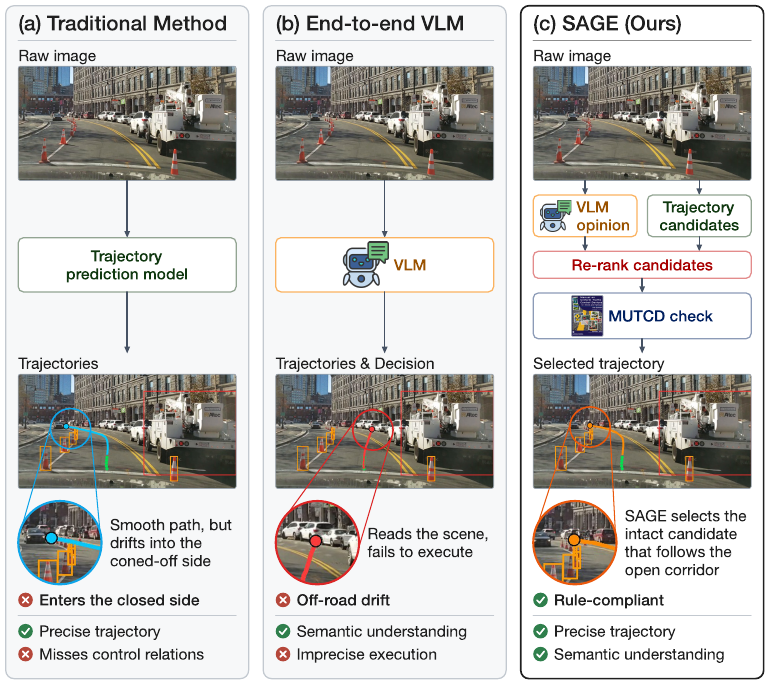}
\caption{Temporary-control-aware pathway prediction. (a) A geometric predictor selects a smooth path toward the coned-off side. (b) A VLM gives a scene reading and an imprecise trajectory. (c) SAGE uses the trajectory as a preference over geometric candidates and the reading as input to MUTCD-derived checks. Paths are illustrative; green dots: observed prefix; orange boxes: cones; red box: work vehicle.}
\label{fig:teaser}
\end{figure}

Roadway work zones temporarily change where and how vehicles may travel. Under temporary traffic control (TTC), signs announce closures, tapers of channelizing devices redirect traffic, and barriers separate the traffic space from the work space. Where the temporary path departs from the permanent alignment, the Manual on Uniform Traffic Control Devices (MUTCD) calls for devices that emphasize the intended path \citep[\S6A.02, \S6B.03--6B.07]{fhwa2025mutcd}. An automated vehicle (AV) must therefore interpret the observed arrangement and translate it into a continuous passing path. The consequences of an incorrect passage can be severe: 850 people died in U.S. work-zone crashes in 2024, including nearly 170 pedestrians and pedalcyclists \citep{fhwa2026nwzaw}. Reliable pathway prediction in this setting requires both an understanding of the temporary control and sufficient spatial precision to pass the workers and devices that define it.

Transportation research has established how work-zone layouts affect capacity and merge management \citep{weng2013capacity,markakis2022merge}, crash occurrence and severity \citep{theofilatos2017meta,osman2018wzseverity}, merging risk and worker protection \citep{weng2015merging,sabeti2024ar}, and drivers' speed, merge timing, and lateral positioning \citep{finley2015speed,weng2016merging,hess2020roadworks,chitturi2025width}. For AVs, these findings motivate a complementary problem: inferring the passing path from the control arrangement visible on board. Work-zone perception methods recognize signs and scenes \citep{seo2015workzone,sundharam2023scenes}, but studies of partially automated driving identify difficulties at temporary markings and cone-delineated lane shifts \citep{naujoks2017limits,cummings2022alerting}. The June 2026 Waymo recall likewise documents vehicles entering construction closures after passing closure signs or between channelizing cones \citep{nhtsa2026recall}. Infrastructure information can assist this interpretation. The Work Zone Data Exchange (WZDx), for example, communicates lane statuses such as closed and shift-left \citep{wzdx2024spec}, and transportation research calls for coordinating such information with vehicle capabilities \citep{dehman2021workzone}. The vehicle must still determine the continuous path through the layout that it encounters.

We study this problem as \emph{temporary-control-aware pathway prediction}: an AV that takes front-view imagery and an observed path prefix as input should predict one future path. A coarse instruction such as \emph{shift left} leaves the onset, extent, and shape of the shift unspecified. Conversely, a smooth continuation of recent motion can miss a redirection indicated by devices farther ahead. ROADWork provides real-world work-zone images, device annotations, scene descriptions, and demonstrated driver pathways from 16 U.S. cities for studying this problem \citep{ghosh2025roadwork}. In more than 90\% of its turning episodes, the demonstrated path crosses our departure threshold only in the far third of the prediction horizon (Section~\ref{sec:transition}). Much of the relevant choice therefore concerns how the path develops beyond its observed prefix.

Geometry-based predictors offer a useful action space for this choice. Waypoint-conditioned models such as PECNet and Y-Net represent multiple plausible futures \citep{mangalam2020pecnet,mangalam2021ynet}, while hierarchical planners generate alternatives and select among them using geometric, kinematic, or cost-based information \citep{yao2025hierarchical,chen2026plancilqr}. In a cone-delineated work-zone lane closure, polynomial trajectory planning similarly constructs alternatives and selects a collision-free path \citep{formosa2024cav}. Device roles and spatial features can inform both generation and ranking, but their contribution depends on whether they help distinguish the passing options in a particular layout. Our candidate-pool analysis locates substantial room for improvement at this selection step: 87.4\% of validation episodes contain a candidate with average displacement error (ADE) below 10\,px, yet selection regret accounts for 53\% of the geometric ranker's error. Geometric selection, however, lacks an interpretation of the temporary control, so in a work zone a smooth and kinematically plausible path can still head toward the coned-off side (Fig.~\ref{fig:teaser}(a)).

Vision--language models (VLMs) provide a complementary source of scene information. Driving-oriented models relate objects and decisions through graph-structured questions \citep{sima2024drivelm}, produce planning decisions or trajectories from camera inputs \citep{you2026v2xvlm,openemma2024,lightemma2025,chi2025impromptu,jiang2024senna}, and supply semantic rewards for driving policies \citep{huang2025vlmrl}. Work-zone studies develop retrieval, memory, latent trajectory reasoning, and causal supervision to improve this capability \citep{liao2025workzonesvlm,luo2025mtrdrive,lu2026onevl,jiang2026workdrive}; complementary work predicts drivable boundaries from synthetic and real work-zone data \citep{sahu2026wzplanner}. Scene interpretation and coordinate prediction nevertheless remain distinct abilities. A model can identify the side to avoid while its path still ends on that side: SAGE-VLM names the described hazard side in 92.7\% of the cases in which it names one, yet 2.25\% of its paths, executed directly, end on that side (Sections~\ref{sec:rq2} and~\ref{sec:z0results}). Such a path can take the vehicle into the closed side or across a channelizing line. Geometry thus provides precise candidates without an interpretation of the temporary control, and a VLM provides the interpretation without a path that reliably follows it; what is missing is a connection between the two.

To bridge this gap, we propose \textbf{SAGE} (Semantic-to-Action Grounded Execution), a read--decide--execute--check framework that connects scene interpretation to precise pathway selection through a shared pool of geometric candidates. Its central insight is that a VLM trajectory can provide a useful preference over candidates even when its own endpoint is less accurate than the geometric prediction. The full trajectory expresses where a passage begins to shift, how far it moves laterally, and how its shape develops, distinguishing alternatives that share the same coarse maneuver. The VLM's structured reading serves a complementary purpose: it identifies the temporary control against which the selected path can be checked. SAGE thus assigns distinct roles to path construction, path preference, and control consistency. Compared with hybrid designs that pass VLM plans or controller parameters to a planner \citep{tian2024drivevlm,long2026vlmmpc} or use learned or VLM-enhanced trajectory scorers \citep{li2024hydramdp,zheng2025simplevsf,huang2026deft}, SAGE preserves geometric coordinates, gives the VLM a fixed contribution to candidate ranking, and applies a separate, bounded correction based on MUTCD-derived checks.

The geometric stream uses a role-decoupled generator, RD-YNet, to construct alternatives from both the visible free space and the temporary-control devices. A learned geometric ranker establishes the default choice. The VLM trajectory then supplies an opinion ranking based on each candidate's agreement with its predicted path, and trajectory-as-opinion (TAO) fusion combines this ranking with the geometric ordering. Semantic guidance therefore changes which path is selected while retaining every candidate's original coordinates. To connect the selected path to explicit control checks, MUTCD-structured supervision trains SAGE-VLM to report the lane status and hazard side before its passage decision and trajectory. The predicted hazard side, together with image-derived device evidence, enters predicates for restricted-space intrusion, channelization crossing, and hazard-side inconsistency. A bounded intervention, Z0, can replace a flagged selection with a candidate that passes these checks, provided that the replacement remains within fixed score-gap and mean-displacement limits. We evaluate TAO and Z0 separately to distinguish improvements in path preference from the effects of corrective checking, using displacement errors, control-consistency measures, and a device-anchored ground frame that estimates lateral offsets in meters.

The main contributions are as follows:
\begin{lightitemize}
\item We propose a shared-candidate framework for temporary-control-aware pathway prediction. SAGE combines a VLM trajectory's preference information with geometric ranking while preserving every candidate's coordinates, and separates this selection interface from the checks informed by the VLM's reading.
\item We develop MUTCD-structured supervision for SAGE-VLM and connect its explicit control reading to a bounded intervention. The model reports the lane status and hazard side before completing its trajectory; the hazard side supplies the interpreted input to predicates on the candidate paths.
\item We establish the value of this division of labor on 2{,}306 ROADWork validation episodes. SAGE reduces ADE and final displacement error (FDE) by 8.3\% and 12.0\% relative to its geometric ranker, and hard-constraint violations by 37\%, and the trajectory opinion improves all four tested generator--selector combinations. SAGE is comparable to OneVL in accuracy with a faster-responding VLM and outperforms four other adapted driving VLMs in displacement error and safety measures.
\item We complement image-space displacement with a device-anchored metric frame, MUTCD-derived consistency measures, and a safety score on human-filtered episodes. These measures link the accuracy analysis to the spatial demands of work-zone passage, and recording-clustered confidence intervals quantify the uncertainty of the gains from the opinion and the intervention.
\end{lightitemize}

Section~\ref{sec:formulation} formulates the problem, and Section~\ref{sec:method} describes SAGE. Section~\ref{sec:experiments} presents the experimental setup, followed by results in Section~\ref{sec:results}. Sections~\ref{sec:discussion} and~\ref{sec:conclusion} discuss the implications and limitations and conclude the paper.

\section{Problem Formulation}
\label{sec:formulation}
We formulate temporary-control-aware pathway prediction as the selection of a continuous path through a visible work zone. The formulation distinguishes the construction of candidate paths from the semantic preferences and control constraints used to select among them.

\subsection{Task definition}
\label{sec:task}
An AV approaching a work zone must infer how the visible temporary traffic control redirects its path. Each prediction instance, termed an \emph{episode}, provides a forward-facing image and an observed path prefix expressed in that image. The task is to predict the continuation of this prefix through the work zone. The recorded human pathway provides the reference for supervised learning and evaluation; its future portion is unavailable to the predictor at inference.

Formally, let $I\in[0,1]^{H\times W\times 3}$ denote an image of height $H$ and width $W$, with image coordinates $p=(u,v)$ representing the column and row. The observed prefix is an ordered sequence $\mathcal O=(o_1,\dots,o_{T_o})$, where $o_t\in\mathbb R^2$. A predictor $\Phi$ maps the image and prefix to a future pathway of $T$ waypoints:
\begin{equation}
\hat{\mathcal F}
=\Phi(I,\mathcal O)
=(\hat f_1,\dots,\hat f_T),
\qquad
\hat f_t\in\mathbb R^2,\quad t=1,\dots,T.
\label{eq:task}
\end{equation}
The reference pathway $\mathcal F=(f_1,\dots,f_T)$ contains the corresponding waypoints from the human demonstration. In ROADWork, the prefix and future are spatially sampled portions of the recorded pathway projected into the source image; they carry no timestamps. The task therefore concerns path geometry, with vehicle speed and response timing outside its scope. Throughout this paper, \emph{execution} refers to selecting an image-space pathway.

We measure agreement with the demonstration using average displacement error (ADE) and final displacement error (FDE):
\begin{equation}
\operatorname{ADE}(\hat{\mathcal F})
=\frac{1}{T}\sum_{t=1}^{T}\|\hat f_t-f_t\|_2,
\qquad
\operatorname{FDE}(\hat{\mathcal F})
=\|\hat f_T-f_T\|_2.
\label{eq:metrics}
\end{equation}
Both are reported in source-image pixels (px). Section~\ref{sec:experiments} specifies the image dimensions and pathway lengths and introduces complementary measures of control consistency and estimated physical displacement.

\subsection{Problem statement}
\label{sec:problem_statement}

Temporary traffic control can redirect a vehicle beyond the continuation suggested by its observed prefix. A single image may show the approach to a work zone, a taper that shifts traffic, and the passage beside the work space. The predicted path must account for these downstream cues while remaining consistent with the visible control arrangement \citep[\S6B.03--6B.08]{fhwa2025mutcd}. Our objective is to combine geometric precision with an interpretation of that arrangement under four design requirements:
\begin{lightitemize}
\item[(D1)] \emph{Accuracy}: the predicted pathway should achieve small ADE and FDE relative to the human demonstration.
\item[(D2)] \emph{Control consistency}: the pathway should respect the visible closure and channelization, as assessed by image-space predicates derived from the applicable MUTCD provisions.
\item[(D3)] \emph{Bounded semantic influence}: the language-based module should influence path selection through an explicit, measurable interface. The output must retain the coordinates of a geometric candidate, and any corrective replacement must remain within specified score and displacement bounds.
\item[(D4)] \emph{Availability}: the system should return a pathway even when the language-based output is unavailable, unparsable, or rejected, using geometric selection as its fallback.
\end{lightitemize}

We address these requirements through three subproblems defined over a shared candidate space:
\begin{lightitemize}
\item[(S1)] \emph{Generation}: construct the available paths. Let $\mathcal R$ denote a set of functional scene roles and $R\in[0,1]^{|\mathcal R|\times H\times W}$ the corresponding role maps predicted from $I$. A geometric generator $F_{\mathrm{RD}}$ uses the image-derived representation and observed prefix to construct $M$ candidate pathways:
\begin{equation}
\mathcal C
=F_{\mathrm{RD}}(I,\mathcal O,R)
=\{C_j\}_{j=1}^{M},
\qquad
C_j=(C_{j,1},\dots,C_{j,T}),
\quad C_{j,t}\in\mathbb R^2.
\label{eq:candidates}
\end{equation}
The pool should cover plausible continuations along the visible road surface and alternatives guided by the temporary devices. SAGE implements this generator with the free-space and TTC-conditioned branches of RD-YNet (Section~\ref{sec:rdynet}).
\item[(S2)] \emph{Preference}: rank the available paths. A geometric scorer assigns each candidate a cost $s_g(j)\in\mathbb R$, with lower values preferred. A trajectory-capable VLM provides an additional prediction,
$\mathcal F^{\mathrm{vlm}}=(f^{\mathrm{vlm}}_1,\dots,f^{\mathrm{vlm}}_T)$,
where $f^{\mathrm{vlm}}_t\in\mathbb R^2$. This trajectory expresses a spatial preference: candidates closer to it receive a stronger VLM preference. SAGE combines this ordering with the geometric ranking through trajectory-as-opinion fusion (Section~\ref{sec:executor}). The VLM thus influences which candidate is selected while preserving its original coordinates.
\item[(S3)] \emph{Compliance}: check the selected path. The VLM also provides a structured interpretation of the temporary control:
\begin{equation}
\mathcal P=(\ell,a_{\mathrm{side}},m),
\qquad
\ell\in\mathcal L,\quad
a_{\mathrm{side}}\in\mathcal A,\quad
m\in\mathcal M,
\label{eq:plan}
\end{equation}
where $\mathcal L$ is a finite set of lane statuses,
$\mathcal A=\{\text{left},\text{right},\text{none or both}\}$
is the set of hazard-side readings, and $\mathcal M$ is a finite set of passage maneuvers. The hazard side identifies the side occupied by the work zone. SAGE-VLM reports the lane status and hazard side before its passage decision and trajectory (Section~\ref{sec:vlm}).

Of these structured fields, the hazard side enters the downstream checks. Together with image-derived device evidence, it supports predicates for restricted-space intrusion, channelization-boundary crossing, and hazard-side inconsistency. If the selected path violates a predicate, a bounded intervention searches the same pool for a replacement that passes the checks and remains close in score and displacement (Section~\ref{sec:tcd_got}). If no admissible replacement exists, the original choice is retained and the violation is recorded as unresolved. These predicates operationalize selected MUTCD provisions within the image-space representation; they do not establish regulatory compliance.
\end{lightitemize}
The three subproblems yield one intact candidate as the final prediction:
\begin{equation}
\hat{\mathcal F}=C_{j^\star},
\qquad j^\star\in\{1,\dots,M\}.
\label{eq:selection}
\end{equation}
Generation determines which paths are available; preference and compliance determine which available path is returned.

\paragraph{Separating generation error from selection regret.}
This shared candidate space allows us to distinguish limitations of the pool from limitations of the selection. Define the \emph{pool floor} as
$E^{\mathrm{pool}}=\min_j\operatorname{ADE}(C_j)$,
the error of the candidate closest to the demonstration. For any selected index $\hat\jmath$, the prediction error decomposes as
\begin{equation}
\operatorname{ADE}(C_{\hat\jmath})
=
\underbrace{E^{\mathrm{pool}}}_{\text{generation}}
+
\underbrace{
\operatorname{ADE}(C_{\hat\jmath})-E^{\mathrm{pool}}
}_{\text{selection regret}}.
\label{eq:regret}
\end{equation}
The pool floor measures the best accuracy available from the generated candidates, while selection regret measures the additional error incurred by the chosen candidate. Both quantities require the reference future and are used only for analysis. A low pool floor combined with high selection regret identifies a specific opportunity for semantic guidance: a suitable path already exists, but the selector fails to choose it.

\section{SAGE: A Read--Decide--Execute--Check Framework}
\label{sec:method}

\begin{figure}[t]
\centering
\includegraphics[width=\linewidth]{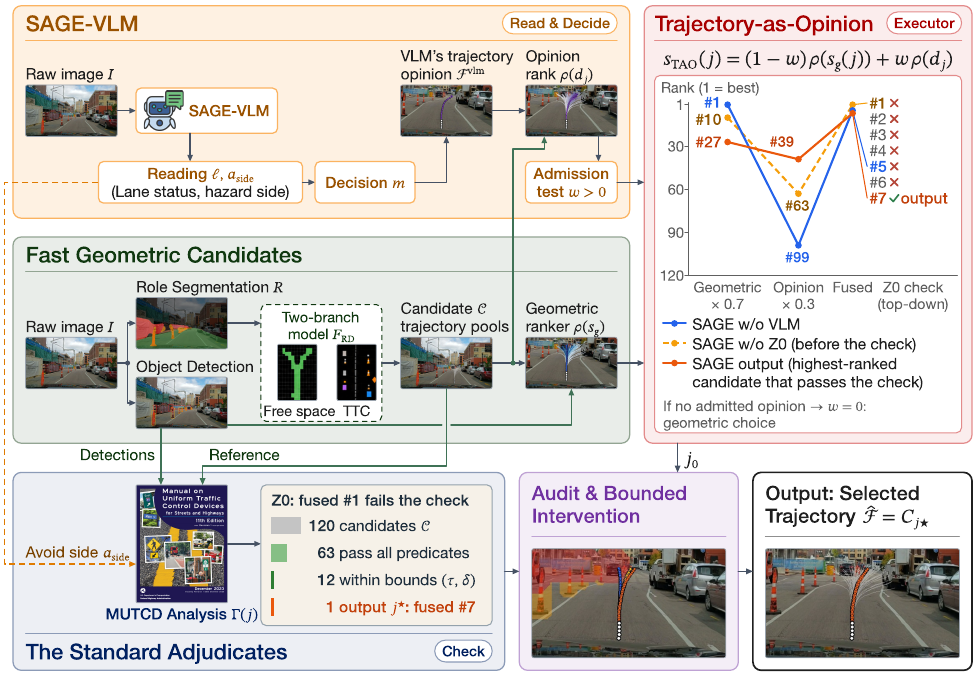}
\caption{Overview of SAGE. Middle band: RD-YNet proposes 60 + 60 candidates, which a geometric ranker scores. Upper band: SAGE-VLM states the lane status, hazard side, passage decision, goal, and trajectory; each candidate's distance to the trajectory gives its opinion rank. Right: rank fusion. Lower band: MUTCD-derived predicates drive the bounded intervention Z0.}
\label{fig:system}
\end{figure}

SAGE connects an interpretation of temporary traffic control to pathway selection through a shared pool of geometric candidates. The framework assigns distinct roles to geometry, language, and explicit constraints: geometry constructs the available paths, the VLM informs their ranking, and MUTCD-derived predicates support a bounded correction of the selected path. This section describes how these components interact while preserving every candidate's original coordinates.

\subsection{Framework overview}
\label{sec:overview}
Fig.~\ref{fig:system} presents the four stages of SAGE. In \emph{read}, SAGE-VLM identifies the lane status and hazard side from the image and observed prefix. In \emph{decide}, it states a passage maneuver and predicts a fifteen-point trajectory. In \emph{execute}, RD-YNet generates geometric candidates, a learned ranker establishes their default ordering, and the VLM trajectory supplies an additional preference over the same pool. In \emph{check}, image-derived evidence and the predicted hazard side enter MUTCD-derived predicates; a bounded intervention can replace a violating choice with a nearby candidate that passes the checks.

These stages define the flow of information. The geometric generator and ranker operate independently of the VLM and can return a pathway before its output is available. Once a usable trajectory from an admitted opinion source is available, trajectory-as-opinion (TAO) fusion refines the ranking. The final intervention, denoted Z0, additionally requires a current reading and a usable trajectory. In all cases, the output is one intact geometric candidate.

Two ablations isolate these contributions. \textbf{SAGE w/o VLM} returns the geometric ranker's choice. \textbf{SAGE w/o Z0} adds TAO fusion but omits the final intervention. Full \textbf{SAGE} includes both. Section~\ref{sec:anchors} also introduces a decision-anchor ablation that compresses the VLM trajectory before using it for selection.

\subsection{Role-decoupled candidate generation}
\label{sec:tcd_representation}
\label{sec:rdynet}
Candidate generation must represent both the continuation suggested by the road surface and the redirection indicated by temporary devices. RD-YNet preserves these alternatives through two branches: one conditioned on the visible surface and one additionally conditioned on device roles. Both branches use the same observed prefix and contribute equally to the candidate pool.

\paragraph{Functional roles.}
A frozen segmentation frontend $f_{\mathrm{seg}}$ maps the image to six aligned role maps:
\[
R=f_{\mathrm{seg}}(I)
\in[0,1]^{|\mathcal R|\times H\times W},
\qquad
\mathcal R=
\{\text{void},\text{drivable},\text{off-road},
\text{hard},\text{channel},\text{guidance}\}.
\]
The frontend consists of a ConvNeXt-Tiny encoder \citep{liu2022convnext} and a progressive-upsampling decoder operating on its stride-32 features. It is trained on the aligned ROADWork segmentation subset before the pathway experiments.

The roles organize scene content by its relevance to passage. The \emph{hard} role includes barriers, barricades, fencing, work and police vehicles, workers, and officers. The \emph{channel} role includes cones, drums, vertical panels, tubular markers, and work equipment. The \emph{guidance} role includes TTC signs and arrow boards. The drivable and off-road maps describe the underlying surface. These groupings support candidate generation: work equipment, for example, contributes to the channel role while remaining an obstacle in the downstream checks.

\paragraph{Role-decoupled construction.}
RD-YNet comprises two Y-Net heatmap predictors \citep{mangalam2021ynet}. Each receives heatmaps of the observed prefix together with its assigned role maps:
\[
\mathcal C_{\mathrm{free}}
=F_{\mathrm{free}}(\mathcal O,R_{\mathrm{surf}}),
\qquad
\mathcal C_{\mathrm{TTC}}
=F_{\mathrm{TTC}}(\mathcal O,R),
\qquad
\mathcal C
=\mathcal C_{\mathrm{free}}\cup\mathcal C_{\mathrm{TTC}},
\]
where $R_{\mathrm{surf}}$ retains the drivable and off-road channels and sets the remaining channels to zero. Thus, the image enters both predictors through the segmentation frontend.

Each branch samples 60 endpoints without replacement from its predicted goal heatmap and decodes one fifteen-point pathway toward each endpoint, yielding $M=120$ candidates. The free-space branch proposes continuations along the visible pavement, while the TTC-conditioned branch can propose shifts and passages shaped by the temporary devices. Neither branch is assigned a maneuver in advance. Equal contributions preserve both sources of alternatives in the final pool.

\begin{figure}[!t]
\centering
\includegraphics[width=0.85\linewidth]{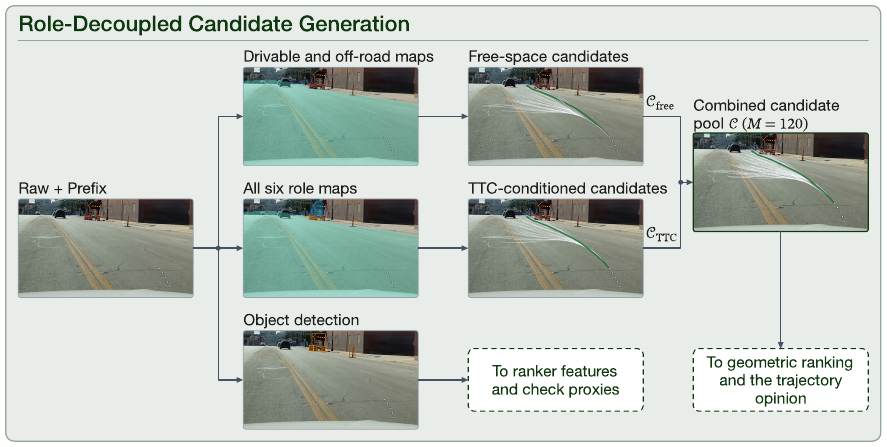}
\caption{Candidate generation with RD-YNet. The free-space branch receives the drivable and off-road maps; the TTC-conditioned branch receives all six role maps. All 10 candidates within 10\,px of the demonstration come from the TTC-conditioned branch. Shading: teal, drivable; gray, off-road; amber, channel; blue, guidance. White lines: candidates; white dots: observed prefix; green: demonstration.}
\label{fig:candidate_generation}
\end{figure}

Fig.~\ref{fig:candidate_generation} illustrates why this separation matters. In the example, all ten candidates within 10\,px of the demonstration come from the TTC-conditioned branch. The pool is fixed before any VLM output is applied; subsequent stages select among its members without modifying their coordinates. Turning feasibility is assessed through the curvature term of the safety score described in Section~\ref{sec:metrics}.

\paragraph{Device evidence.}
The geometric ranker and the constraint checks also use the TTC object detections released with each ROADWork episode \citep{ghosh2025roadwork}. For ranking, side-specific counts summarize the hard and channelization detections:
\begin{equation}
G=(h_L,h_R,q_L,q_R),
\label{eq:graph}
\end{equation}
where $h_L,h_R$ count hard-role boxes on the left and right, and $q_L,q_R$ count channelization boxes. Candidate-to-device distances provide additional spatial information, while the checks use the detection boxes themselves.

\subsection{Reading and deciding with SAGE-VLM}
\label{sec:vlm}

SAGE-VLM supplies two complementary inputs to selection: a trajectory that expresses a preferred passage and a hazard-side reading that supports explicit checks. Its output format places the interpretation before the trajectory, making the reading available earlier and conditioning subsequent tokens on it.

\paragraph{Model adaptation.}
The released ROADWork checkpoint of OneVL \citep{lu2026onevl} is trained to produce trajectories with latent reasoning tokens. When prompted to state the lane status and hazard side before its path, it continues to emit only the trajectory. Its auxiliary language decoder can translate the latent reasoning into text, but this takes approximately 30\,s per frame under the evaluated configuration (Section~\ref{sec:efficiency}). SAGE therefore trains the reading as an explicit part of the main model's output.

SAGE-VLM is initialized from the main Qwen3-VL-4B network \citep{bai2025qwen3vl} of the OneVL ROADWork checkpoint, without its auxiliary latent decoders. We fine-tune low-rank adapters \citep[LoRA;][]{hu2022lora} on all language-model linear layers and the vision tower.

\paragraph{MUTCD-structured supervision.}
The input consists of the image and five observed waypoints represented on a 0--1000 coordinate grid, denoted $\tilde{\mathcal O}$. The model emits a text sequence in the following order:
\begin{quote}\small
\texttt{Work zone: lane status=}$\ell$\texttt{; hazard side=}$a_{\mathrm{side}}$\\
\texttt{Scene:} a short description\\
\texttt{Decision:} $m$\\
\texttt{Goal: (}$x,y$\texttt{)}\\
\texttt{Path:} fifteen points
\end{quote}
Parsing this sequence yields
\[
(\ell,a_{\mathrm{side}},m,\tilde{\mathcal F}^{\mathrm{vlm}})
=\pi_{\mathrm{VLM}}(I,\tilde{\mathcal O}),
\]
where
$\mathcal L=\{\text{lane closed},\text{road closed},
\text{lane shifted},\text{open}\}$,
$\mathcal A=\{\text{left},\text{right},\text{none or both}\}$,
and
$\mathcal M=\{\text{continue straight},\text{shift left},
\text{shift right},\text{turn left},\text{turn right}\}$.

The supervision organizes ROADWork annotations around concepts used in temporary traffic control. Lane-status targets follow the scene-level travel-alteration tags, including partially blocked, fully blocked, and lane shift. The scene sentence is the human description of the frame. A fixed rule extracts the hazard side from clauses that explicitly name a side of the road.

The passage decision and trajectory targets come from the demonstrated pathway. An endpoint displacement beyond $0.04W$ from the current position determines the lateral direction, and a heading change above $30^\circ$ distinguishes a turn from a shift. The goal is the final demonstrated waypoint. This sequence trains the model to state the control reading before producing the decision and path.

\paragraph{Output conversion.}
The predicted trajectory is converted from the 0--1000 grid to source-image pixels:
\begin{equation}
f^{\mathrm{vlm}}_t
=(1.92\,\tilde x_t,\;1.08\,\tilde y_t),
\qquad t=1,\dots,T.
\label{eq:conversion}
\end{equation}
All subsequent path comparisons use these source-image coordinates. Only two outputs enter the downstream selection: $\mathcal F^{\mathrm{vlm}}$ supplies the trajectory opinion, and $a_{\mathrm{side}}$ supplies the interpreted input to the checks. The remaining fields make the model's reading and passage decision observable.

\subsection{Candidate selection with geometry and trajectory opinion}
\label{sec:executor}
\label{sec:fusion}

TAO combines two orderings of the same candidates. The geometric ranking reflects path and device features, while the opinion ranking reflects agreement with the VLM's predicted passage. Combining them in rank space gives the VLM a fixed contribution without requiring the two scores to share a numerical scale.

\paragraph{Geometric ranking.}
Each candidate $C_j$ is represented by a feature vector $\phi_j$ containing its branch identity, heatmap score and rank, endpoint lateral offset, path length, curvature, deviation from observed-motion extrapolation, the device counts $G$, and the nearest hard and channelization distances on each side.

A HistGradientBoosting regressor $h_\theta$ \citep{pedregosa2011sklearn} predicts the candidate's within-episode percentile rank of ADE:
\[
s_g(j)=h_\theta(\phi_j),
\qquad
j_g=\arg\min_j s_g(j).
\]
The regressor uses maximum depth 4, learning rate 0.04, L2 regularization 1, and at most 500 iterations with scikit-learn's default early stopping. Reference-path ADE supplies training labels only. At inference, the ranker uses observable candidate and device features.

The selected candidate $C_{j_g}$ is the output of SAGE w/o VLM and serves as the fallback when no admitted opinion is available. Because this ranker already uses device evidence, the VLM's contribution is evaluated beyond a geometric baseline that includes work-zone information.

\paragraph{Trajectory opinion.}
A coarse maneuver can assign the same label to paths with different shift onsets, lateral extents, and shapes. The complete VLM trajectory preserves these distinctions. We quantify its agreement with candidate $C_j$ by the mean pointwise distance:
\begin{equation}
d_j
=D(C_j,\mathcal F^{\mathrm{vlm}})
=\frac{1}{T}\sum_{t=1}^{T}
\|C_{j,t}-f^{\mathrm{vlm}}_t\|_2,
\qquad j=1,\dots,M.
\label{eq:opinion_distance}
\end{equation}
Smaller distances imply stronger preference. Any trajectory-capable model following the coordinate convention in Section~\ref{sec:vlm} can supply this opinion; full SAGE uses SAGE-VLM.

\paragraph{Rank-space fusion.}
Let $\rho(\cdot)$ convert scores to normalized within-pool ranks. A rank $r\in\{1,\dots,M\}$ becomes $(r-1)/(M-1)$, with ties assigned their average rank. Lower values are preferred. The fused score and selected index are
\begin{equation}
\begin{aligned}
s(j)
&=
\begin{cases}
s_{\mathrm{TAO}}(j)
=(1-w)\rho(s_g(j))+w\rho(d_j),
& \text{if an admitted opinion is available},\\
\rho(s_g(j)),
& \text{otherwise},
\end{cases}\\
j_0&=\arg\min_j s(j),
\end{aligned}
\label{eq:fusion}
\end{equation}
with $w=0.3$. The candidate $C_{j_0}$ is the output of SAGE w/o Z0.

Rank normalization prevents differences in the scales of learned costs and pixel distances from determining their relative contribution. The fixed weight gives the geometric ranking greater weight while allowing the opinion to change the selected candidate. Simple fixed combinations also have an established motivation in forecasting, where predictors with different errors can improve a combined estimate \citep{bates1969combination,claeskens2016puzzle}. Here, fusion changes only the selected index; all candidate coordinates remain unchanged.

\paragraph{Opinion-source admission.}
An opinion source is used only if it demonstrates value on held-out training data. We evaluate $w\in\{0,0.1,\dots,0.5\}$ on a 339-episode development subset excluded from SAGE-VLM fine-tuning, using snippet-grouped out-of-fold geometric predictions. A source whose optimum is $w=0$ is rejected and contributes no opinion ranking. For admitted sources, the fusion weight is fixed at 0.3. SAGE-VLM passes this admission test.

Admission is a source-level decision. If an admitted source provides no usable trajectory for a particular episode, selection falls back to the geometric ranking.

\paragraph{Decision anchors.}
\label{sec:anchors}
To test whether the complete trajectory is needed, we construct an ablation that retains only a small set of decision anchors. Douglas--Peucker simplification \citep{douglas1973algorithms}, with tolerance 15\,px, is applied to the polyline from the last observed waypoint through $\mathcal F^{\mathrm{vlm}}$. The anchors retain the goal and at most two intermediate key points. If the path cannot be parsed but the goal can, the goal alone is used.

For this ablation, $d_j$ is replaced by the mean of the goal-to-endpoint distance and the intermediate-anchor-to-candidate distances; the rank fusion remains unchanged. This interface reduces the trajectory information passed to the selector, allowing us to assess the value of the full path shape (Section~\ref{sec:sources}).

\subsection{Standard-grounded checks and bounded intervention}
\label{sec:tcd_got}
\label{sec:vlm_planner}

A high-ranked candidate can still conflict with the visible control. The final stage therefore checks the fused choice against explicit predicates and, when possible, replaces it with a nearby candidate that passes them. Its inputs are the candidate pool, the fused scores, the role maps and detections, and the VLM's hazard-side reading.

\paragraph{MUTCD provisions.}
Table~\ref{tab:mutcd_predicates} links the predicates to MUTCD provisions \citep{fhwa2025mutcd}. Image-derived regions and detections provide spatial evidence, while the VLM reading identifies the hazard side. The predicates use this information to assess whether a candidate enters the work space, crosses the channelization proxy, or ends on the interpreted hazard side.

The checks cover provisions that can be represented with the available image-space paths and scene evidence. Buffer and taper lengths, lateral buffer widths, and flagger control require additional depth, speed, timing, or engineering judgments and are outside the present formulation (\S6B.06, \S6B.08, \S6D.05). Accordingly, the checks are operational proxies for selected provisions; establishing regulatory compliance requires an engineering assessment.

\begin{table}[pos=!t]
\centering
\caption{MUTCD provisions instantiated as image-space predicates of the count $\Gamma$ in Eq.~\eqref{eq:gamma}. RSI: restricted-space intrusion; CBC: channelization-boundary crossing.}
\label{tab:mutcd_predicates}
\footnotesize
\renewcommand{\arraystretch}{1.12}
\begin{tabularx}{\linewidth}{@{}>{\raggedright\arraybackslash}p{2.7cm}>{\raggedright\arraybackslash}X>{\raggedright\arraybackslash}p{3.3cm}>{\raggedright\arraybackslash}p{3.1cm}@{}}
\toprule
Provision & Requirement (paraphrased) & Predicate on $C_j$ & Evidence \\
\midrule
\S6B.06 \P02 & The work space is closed to road users. & $\mathrm{RSI}(C_j)>0$ & Hard-role regions $\cup$ hard and activity detections \\
\S6B.05 \P02 & Redirected road users are guided to a new path by channelizing devices. & $\mathrm{CBC}(C_j)>0$ & Channel-role regions $\cup$ channelizing-device detections \\
\S6B.05 \P02; \S6B.06 \P02; \S6L.06 \P17, \P19 & Traffic passes beside the closed side; arrow boards mark lane closures. & $\mathrm{side}(C_j)=a_{\mathrm{side}}$ & SAGE-VLM's reading \\
\bottomrule
\end{tabularx}

\end{table}

\paragraph{Predicates.}
We evaluate three quantities for each candidate:
\begin{itemize}[leftmargin=1.2em,itemsep=1pt,topsep=2pt]
\item \emph{Restricted-space intrusion}, $\mathrm{RSI}(C_j)$, is the fraction of the densified path inside a proxy combining hard-role regions with hard-device and work-activity detection boxes dilated by $0.02W$. These detections include barriers, barricades, fences, workers, police officers, work and police vehicles, and work equipment. The proxy represents the work space reserved for workers, equipment, and materials (\S6B.06).
\item \emph{Channelization-boundary crossing}, $\mathrm{CBC}(C_j)$, counts entries of the densified path into a proxy combining channel-role regions with channelizing-device and arrow-board boxes dilated by approximately $0.01W$. It represents the channelization that guides redirected traffic (\S6B.05).
\item \emph{Endpoint side}, $\mathrm{side}(C_j)\in\{\text{left},\text{right},\text{neutral}\}$, is determined by the signed lateral offset of the endpoint from the line through $o_{T_o}$ along the observed heading, with a neutral band of $0.04W$.
\end{itemize}

The number of violated predicates is
\begin{equation}
\Gamma(j)
=
\mathbf1[\mathrm{side}(C_j)=a_{\mathrm{side}}]
+\mathbf1[\mathrm{RSI}(C_j)>0]
+\mathbf1[\mathrm{CBC}(C_j)>0].
\label{eq:gamma}
\end{equation}
A hazard-side reading of ``none or both'' contributes zero to the first term. The reading affects these predicates but does not alter the fused score $s(j)$.

\paragraph{Bounded intervention (Z0).}
\label{sec:z0}
Z0 runs when a current reading and a usable trajectory are available. If $\Gamma(j_0)=0$, it retains the fused choice. Otherwise, it searches for candidates that pass all three predicates and remain close to that choice in both score and mean displacement:
\begin{equation}
\mathcal J
=
\left\{
j:
\Gamma(j)=0,\;
s(j)-s(j_0)\le\tau,\;
D(C_j,C_{j_0})\le\delta
\right\}.
\label{eq:veto_set}
\end{equation}
Here, $\tau=0.15$ bounds the score gap $\Delta s=s(j)-s(j_0)$ on the normalized-rank scale, and $\delta=0.02W=38.4$\,px bounds the mean pointwise displacement. On a single ranking of 120 candidates, a normalized gap of 0.15 corresponds to approximately 18 rank positions.

The final selection is
\begin{equation}
j^\star
=
\begin{cases}
\displaystyle\arg\min_{j\in\mathcal J}s(j),
& \Gamma(j_0)>0\ \text{and}\ \mathcal J\neq\varnothing,\\
j_0,
& \text{otherwise},
\end{cases}
\qquad
\hat{\mathcal F}=C_{j^\star}.
\label{eq:z0_output}
\end{equation}
Thus, Z0 selects the highest-ranked admissible replacement. If the fused choice violates a predicate but no replacement satisfies the bounds, the original choice is retained and the outcome is recorded as unresolved.

The predicates contain no learned parameters. The displacement bound and proxy conventions are fixed a priori, while $\tau$ is selected on the training split. The two stages impose different restrictions: TAO fixes the opinion's contribution to the ranking, and Z0 bounds its corrective replacement in score and path space. Algorithm~\ref{alg:sage} summarizes the complete inference procedure.

\begin{algorithm}[!t]
\caption{Candidate-grounded SAGE inference.}
\label{alg:sage}
\begin{algorithmic}[1]
\Require Image $I$, observed prefix $\mathcal O$, frozen geometric modules, optional current VLM output
\State Predict role maps $R$ and summarize detected TTC evidence $G$
\State Generate $\mathcal C_{\mathrm{free}}$ and $\mathcal C_{\mathrm{TTC}}$; form the 60 + 60 candidate pool
\State Score candidates with $s_g$; set $s\gets\rho(s_g)$ and $j_0\gets\arg\min_j s(j)$
\If{a usable current trajectory from an admitted source is available}
\State Convert its coordinates analytically and compute $d_j$ from Eq.~\eqref{eq:opinion_distance}
\State Set $s\gets s_{\mathrm{TAO}}$ and $j_0\gets\arg\min_j s(j)$ using Eq.~\eqref{eq:fusion}
\EndIf
\State Set $j^\star\gets j_0$
\If{Z0 is enabled and a current reading with a usable trajectory is available}
\State Construct $\mathcal J$ in Eq.~\eqref{eq:veto_set} with $a_{\mathrm{side}}$ from the reading
\If{$\Gamma(j_0)>0$ and $\mathcal J\ne\varnothing$}
\State $j^\star\gets\arg\min_{j\in\mathcal J}s(j)$
\EndIf
\EndIf
\State \Return $C_{j^\star}$, decision source, reading, and audit record
\end{algorithmic}
\end{algorithm}

\begin{figure}[!t]
\centering
\includegraphics[width=0.96\linewidth]{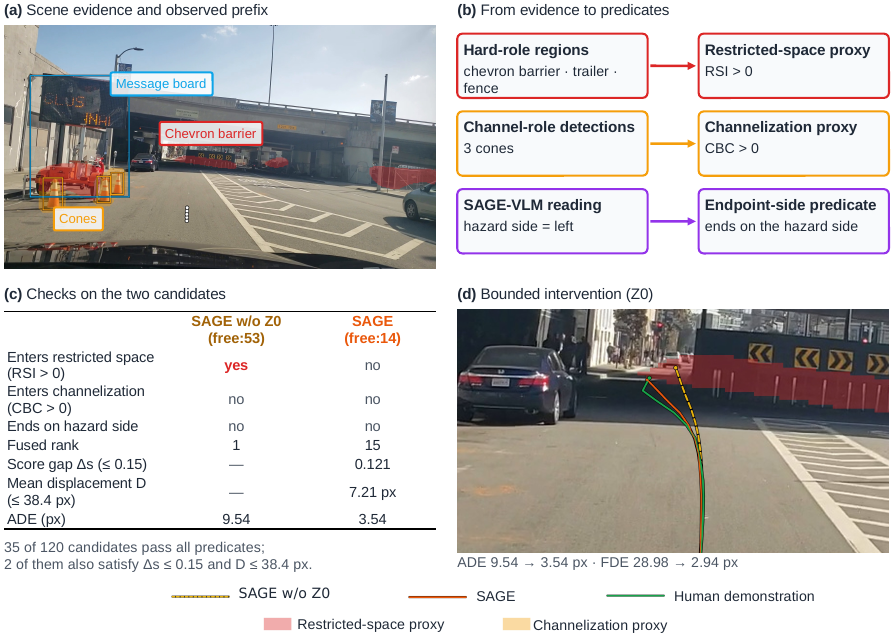}
\caption{From scene evidence to a checked pathway. (a) Detections and observed prefix. (b) Author-constructed summary linking role regions, detections, and the VLM's hazard-side reading to the predicates. (c) Checks on the choices of SAGE w/o Z0 and SAGE. (d) Z0 replaces the SAGE w/o Z0 choice, which enters the restricted-space proxy, with an admissible candidate. Green: demonstration, used only for evaluation.}
\label{fig:semantic_to_path}
\end{figure}

\paragraph{Grounded example.}
Fig.~\ref{fig:semantic_to_path} illustrates the intervention in a San Francisco work zone. The fused choice enters the restricted-space proxy at a barrier ahead. Of the 120 candidates, 35 pass the predicates, and two also satisfy the score and displacement bounds. Z0 selects the higher-ranked of these two, yielding a replacement 7.21\,px from the original choice while preserving its geometric coordinates. The example links the visible evidence and hazard-side reading to a specific, bounded change in the selected pathway.

\section{Experimental Setup}
\label{sec:experiments}
We evaluate whether SAGE improves pathway selection, how its trajectory opinion and bounded intervention contribute to that improvement, and what computational cost they introduce. The evaluation combines displacement errors with control-consistency checks, approximate physical-unit measures, and a human review of the changed paths.

\subsection{Dataset}
\label{sec:data}
We use 3{,}117 training and 2{,}313 validation pathway episodes from ROADWork \citep{ghosh2025roadwork}, which are drawn from real driving videos recorded in 16 U.S. cities. The scenes range from lane shifts and roads partially blocked by work vehicles to activity areas with workers on the road or sidewalk. Each episode provides $T_o=5$ observed and $T=15$ future waypoints in $W\times H=1920\times1080$ source images, and the trajectory networks operate on a $448\times256$ working grid, whose displacements are scaled by $\operatorname{diag}(1920/448,\,1080/256)$ back to source-image pixels before evaluation. ROADWork projects the path that the vehicle drove after the image was taken into the image and samples it at equal spacing along the path, without timestamps; the observed prefix is therefore the first stretch of this path and begins where the image was taken. VLMs are additionally trained on ROADWork's dense training trajectories, which sample the training videos at 5\,Hz. Because ROADWork releases no test pathways, the official validation split serves as the test set. We evaluate on the 2{,}306 validation episodes for which OneVL, the strongest external baseline, returns a parsable trajectory (all but seven).

\subsection{Evaluation metrics}
\label{sec:metrics}
\paragraph{Displacement.}
ADE and FDE (Eq.~\eqref{eq:metrics}) are the primary metrics. To locate where a method gains, we also average the displacement over the near (waypoints 1--5), middle (6--10), and far (11--15) segments.

\paragraph{Diagnostic measures.}
\label{sec:transition}
Two diagnostics locate the selection difficulty; they use the demonstrated future and therefore serve only for analysis. The first measures how a path departs from the continuation of recent motion. With $\bar f_t\in\mathbb R^2$ the cubic extrapolation of the observed prefix and $\hat n\in\mathbb R^2$ the unit normal to the observed heading at $o_{T_o}$, pointing to the right of the image, the signed lateral residual at $t\in\{1,\dots,T\}$ is
\begin{equation}
r_t=\langle f_t-\bar f_t,\hat n\rangle .
\label{eq:residual}
\end{equation}
A path is \emph{turning} if $|r_T|$ exceeds the lateral band $b=0.04W=76.8$\,px, and its departure onset is the first $t$ at which $|r_t|$ and $|r_{t+1}|$ both exceed $\max(38.4\,\mathrm{px},\,0.35|r_T|)$. Of the 2{,}313 validation episodes, 25.7\% are turning, and in 90.9\% of these the onset falls in the far segment. An episode is \emph{laterally departing} if its endpoint $f_T$ lies farther than $b$ from the line of the observed heading through $o_{T_o}$. The second diagnostic separates the candidate pool from the selection with the decomposition of Eq.~\eqref{eq:regret}: a low pool floor $E^{\mathrm{pool}}$ with a high selection regret calls for better selection.

\paragraph{Device-anchored metric frame.}
ROADWork's pathway annotations carry no per-frame calibration, so displacement is commonly reported in source-image pixels \citep{ghosh2025roadwork,liao2025workzonesvlm,jiang2026workdrive}. To express paths in traffic-engineering units, we anchor a flat-ground pinhole model to MUTCD drums. An object of height $H_{\mathrm{obj}}$ whose ground contact lies at image row $v$ spans $H_{\mathrm{obj}}(v-v_h)/h_c$ pixels, with horizon row $v_h$ and camera height $h_c$ \citep{hoiem2008perspective}, and a drum is at least 36\,in (0.91\,m) high \citep[\S6K.06]{fhwa2025mutcd}. Each recording with at least eight eligible drums receives a least-squares fit of $(v_h,h_c)$, which covers 996 of the 2{,}306 evaluation episodes, and the other recordings use the median fit. A ground point at pixel $(u,v)$ then lies at the lateral offset $X=(u-W/2)h_c/(v-v_h)$ and the distance $Z=f_{\mathrm c}h_c/(v-v_h)$, with the focal length $f_{\mathrm c}=2{,}059$\,px for ROADWork's field of view of about $50^\circ$ \citep{ghosh2025roadwork}, and a detection box touches the ground at its bottom row \citep{jiang2026workdrive}. In this frame, we report
\begin{lightitemize}
\item the \emph{row-matched lateral error} (m), in rows at least 80\,px below the horizon, roughly the nearest 30\,m;
\item the \emph{lane-level miss rate} (\%), the share of paths whose maximum lateral error exceeds 1.83\,m, half of a 12-ft lane \citep{aashto2018greenbook};
\item the \emph{collision rate} (\%), the share of paths whose footprint, with the 1.07-m half-width of the passenger car of the American Association of State Highway and Transportation Officials (AASHTO), overlaps a work-zone object (worker, work vehicle, equipment, channelizing device, or barrier) by more than 0.3\,m, the tolerance that nuPlan allows for drivable-area violations \citep{caesar2021nuplan}.
\end{lightitemize}

\paragraph{Safety score.}
Following NAVSIM's predictive-driver-model (PDM) score \citep{dauner2024navsim,cao2025pseudosim}, we use $S=\mathrm{NC}\times\mathrm{DAC}\times\mathrm{CLC}\times(5\,\mathrm{EP}+2\,\mathrm{CF})/7$ on its 0--100 scale. No collision (NC) requires that the path strike no object; drivable-area compliance (DAC) requires that at most 20\% of the path lie in the restricted-space proxy of the work space; and channelization compliance (CLC), a work-zone term in the manner of NAVSIM v2, requires that the path stay out of the channelization proxy. Ego progress (EP) is the progress along the observed heading relative to the demonstration, capped at 1, and curvature feasibility (CF), which takes the place of NAVSIM's comfort term, requires a ground-plane radius of at least 7.3\,m, the minimum turning radius of the AASHTO passenger car \citep{aashto2018greenbook}; NAVSIM's time-to-collision term is omitted because paths carry no timing. A path that fails NC, DAC, or CLC has a \emph{hard violation} and scores zero. Because recorded human paths also pass close to objects, the collision rate, hard violations, and the score use the 2{,}086 \emph{human-filtered} episodes whose demonstration passes all three terms; NAVSIM v2 applies such a filter term by term. We also report hard violations and the score on the 137 \emph{safety-critical frames}, the human-filtered episodes in which constant-velocity extrapolation commits a hard violation.

\paragraph{Control consistency.}
\label{sec:metrics_consistency}
The hazard-side share is the share of paths that end beyond the $0.04W$ neutral band on the side named in the human scene description. Of the 1{,}474 episodes whose description names a side, 12.1\% of the demonstrations themselves end on that side, for instance when they return to their lane, so we compute it on the 1{,}295 described-side episodes whose demonstration ends elsewhere; the description rule is the one that defines SAGE-VLM's training targets, applied to validation descriptions. We also report restricted-space intrusion ($\mathrm{RSI}>0$), channelization-boundary crossing ($\mathrm{CBC}>0$), and the share of choices flagged by $\Gamma$. We read these measures together with displacement, because a short or displaced path can score well on them.

\paragraph{Uncertainty and human review.}
Intervals are 95\% percentile intervals from 10{,}000 bootstrap resamples of the 187 recordings, because episodes of one recording share scene content. Differences between shares are given in percentage points (pp). In addition, one annotator reviews all 93 episodes in which Z0 changes the path, each paired with an unchanged episode from the same recording. The annotator sees the paths before and after the check in random A/B order, with method names, masks, boxes, and the demonstration hidden, and judges whether each path is consistent with the visible closure or channelized passage and whether it clearly contacts a worker or device.

\subsection{Baselines}
\label{sec:baselines}
All methods receive the same image and observed prefix. The baselines form three groups, each of which tests one part of the design.
\begin{lightitemize}
\item \emph{Path models without language} show what the observed motion and geometry achieve alone. Constant-velocity extrapolation measures how much of the path follows from recent motion; the generic planner executes the highest-scoring of 100 samples of the free-space RD-YNet branch alone, without device roles or ranker; and ROADWork's released Y-Net pathway model, run as in its official notebook with one goal and one sample \citep{ghosh2025roadwork}, is the dataset's own reference.
\item \emph{VLM/VLA paths executed directly} test the alternative to the SAGE interface, in which the VLM's trajectory itself is the output. OneVL \citep{lu2026onevl} is a driving VLA released with a ROADWork checkpoint, whose network SAGE-VLM starts from. LightEMMA, OpenEMMA, Impromptu-VLA, and Senna \citep{lightemma2025,openemma2024,chi2025impromptu,jiang2024senna} represent open driving VLMs and VLAs, which we adapt to ROADWork with LoRA; the prompts of LightEMMA and OpenEMMA also carry ROADWork's human scene description, so these two serve as privileged references. A zero-shot Qwen2.5-VL-7B with OneVL's prompt \citep{qwen2025vl} shows what a general-purpose VLM provides, and SAGE-VLM executed directly isolates the effect of the interface on the same trajectory.
\item \emph{SAGE and its ablations} isolate each component on one shared candidate pool (Table~\ref{tab:sage_variants}): SAGE w/o VLM measures the geometric layer, SAGE w/o Z0 adds the trajectory opinion, and SAGE adds the bounded intervention. To compare the cost of obtaining a reading, we also decode OneVL's latent reasoning with its auxiliary language decoder.
\end{lightitemize}
The adapted baselines are task-aligned comparisons whose scores refer to our inputs and adaptation protocol. Direct-output metrics use each method's valid outputs, which cover at least 99.3\% of the episodes, and an invalid response is never replaced by a SAGE prediction.

\begin{table}[pos=!tb]
\centering
\caption{Comparison of SAGE variants and their components.}
\label{tab:sage_variants}
\footnotesize
\begin{tabular}{@{}lccccc@{}}
\toprule
Variant
& \makecell{RD-YNet\\candidates}
& \makecell{Geometric\\ranker}
& \makecell{VLM guidance for\\candidate selection (TAO)}
& \makecell{VLM reading\\+ checks (Z0)}
& Output \\
\midrule
SAGE-VLM, executed directly
& \xmark & \xmark & \xmark & \xmark
& $\mathcal F^{\mathrm{vlm}}$ \\
SAGE w/o VLM
& \cmark & \cmark & \xmark & \xmark
& $C_{j_g}$ \\
SAGE w/o Z0
& \cmark & \cmark & \cmark & \xmark
& $C_{j_0}$ \\
SAGE w/o Z0, decision anchors
& \cmark & \cmark & \makecell{anchors\\only} & \xmark
& $C_{j_0}$ \\
\rowcolor{sagepale}
SAGE (ours)
& \cmark & \cmark & \cmark & \cmark
& $C_{j^\star}$ \\
\bottomrule
\end{tabular}
\end{table}

\subsection{Implementation details}
\label{sec:implementation}
The two RD-YNet branches are trained on all 3{,}117 training episodes for 20 epochs, and we keep the final epoch. To give the ranker unseen training candidates, we regenerate the training-split candidates with five generators, each trained on four of five folds grouped by recording, and fit the ranker on these cross-fitted candidates. SAGE-VLM is fine-tuned with LoRA rank 64, learning rate $10^{-4}$ with cosine decay, gradient accumulation 16, and three epochs on 17{,}260 training examples, namely the training-split episodes and the dense training frames. The video snippets of a 339-episode development subset are held out from SAGE-VLM's fine-tuning and used for epoch selection and the admission test, and the third epoch has the lowest development path ADE. Images are resized to at most $1{,}048{,}576$ pixels, and decoding is greedy with at most 384 new tokens. Each frame is evaluated with the VLM output computed on that frame, which corresponds to an opinion that is current when the geometric choice is made; in operation, the reading arrives after about 0.7\,s and the complete output after about 3\,s (Section~\ref{sec:efficiency}), and the geometric layer does not wait for either. We measure all timing results with batch size 1 on one RTX 6000 Ada GPU of a two-GPU workstation, with no other process on that GPU; the comparison of SAGE-VLM with OneVL runs both in the same code, one after the other, under the same idle conditions.

\section{Results}
\label{sec:results}
Section~\ref{sec:rq2} compares SAGE with the baselines. Section~\ref{sec:sources} then analyzes where the gain arises, Section~\ref{sec:z0results} how the reading supports the bounded intervention, and Section~\ref{sec:efficiency} how fast the framework responds. Section~\ref{sec:qualitative} closes with qualitative cases.

\subsection{Main results}
\label{sec:rq2}
\begin{figure}[!t]
\centering
\includegraphics[width=0.96\linewidth]{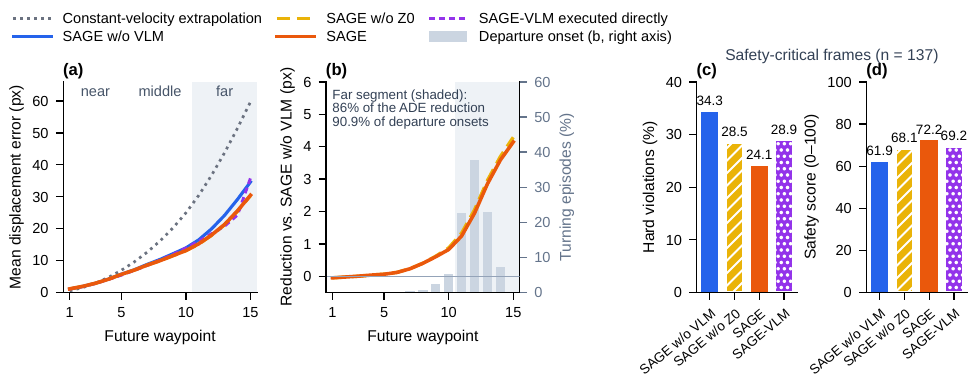}
\caption{Error along the horizon and safety. (a) Mean displacement error per future waypoint. (b) Lines: error reduction relative to SAGE w/o VLM; bars: departure onset of the 594 turning episodes. (c, d) Hard-violation rate and safety score on the 137 safety-critical frames.}
\label{fig:horizon}
\end{figure}

Table~\ref{tab:main_results} shows that SAGE has the best safety measures of the compared methods, with an accuracy at the level of the strongest baseline. Relative to SAGE w/o VLM, it lowers ADE by 8.3\% and FDE by 12.0\%, reduces the share of paths with a hard violation by 37\%, from 2.73\% to 1.73\%, and halves the share of paths that end on the hazard side. Because SAGE and its ablations select from the same candidate pool, these gains come from selection. The trajectory opinion provides the accuracy gain and about three-fifths of the reduction in hard violations, and the bounded intervention adds the rest, as Section~\ref{sec:z0results} shows. Since hard violations are rare in most episodes, the safety gain is clearest on the 137 safety-critical frames of Fig.~\ref{fig:horizon}(c,~d), where SAGE lowers the hard-violation rate from 34.3\% to 24.1\%. The opinion and the check thus matter most where the temporary control requires the vehicle to depart from its current motion.

\begin{table*}[pos=!t]
\centering
\caption{Validation results on 2{,}306 episodes. Collision, hard violations, and score use the 2{,}086 human-filtered episodes; hazard side uses the 1{,}295 described-side episodes. Bold: best; underline: second best; --: not evaluated.}
\label{tab:main_results}
\footnotesize
\setlength{\tabcolsep}{2.6pt}
\begin{threeparttable}
\begin{tabular}{@{}lcccccccc@{}}
\toprule
 & \multicolumn{2}{c}{Displacement (px)} & \multicolumn{2}{c}{Metric frame} & \multicolumn{3}{c}{Safety} & Control \\
\cmidrule(lr){2-3}\cmidrule(lr){4-5}\cmidrule(lr){6-8}\cmidrule(l){9-9}
Method & ADE\,$\downarrow$ & FDE\,$\downarrow$ & \makecell{Lat.\ err.\\(m)\,$\downarrow$} & \makecell{Lane miss\\(\%)\,$\downarrow$} & \makecell{Collision\\(\%)\,$\downarrow$} & \makecell{Hard viol.\\(\%)\,$\downarrow$} & \makecell{Score\\(0--100)\,$\uparrow$} & \makecell{Hazard side\\(\%)\,$\downarrow$} \\
\midrule
Constant velocity & 21.51 & 59.76 & 0.174 & 6.68 & 2.06 & 6.57 & 93.1 & -- \\
Generic planner & 14.06 & 35.08 & 0.133 & 3.78 & 1.01 & 2.40 & 95.9 & 1.16 \\
ROADWork Y-Net & 38.58 & 75.99 & 0.220 & 6.38 & 1.49 & 3.45 & 86.9 & 4.17 \\
\midrule
Qwen2.5-VL-7B & 25.81 & 64.90 & 0.196 & 6.99 & 2.01 & 6.66 & 91.8 & \textbf{0.54} \\
Senna & 18.73 & 54.24 & 0.145 & 5.54 & 1.88 & 6.36 & 93.3 & 1.24 \\
Impromptu-VLA & 17.26 & 52.24 & 0.126 & 4.51 & 1.44 & 5.18 & 94.2 & 1.85 \\
LightEMMA$^\dagger$ & 14.42 & 47.55 & \textbf{0.104} & 4.21 & 1.34 & 5.08 & 94.5 & 2.78 \\
OpenEMMA$^\dagger$ & 14.45 & 47.50 & \second{0.107} & 4.35 & 1.15 & 5.05 & 94.6 & 2.24 \\
OneVL & 12.74 & \textbf{29.02} & 0.114 & 3.26 & \textbf{0.77} & 2.83 & 96.5 & 2.63 \\
SAGE-VLM & 12.15 & 35.86 & 0.110 & \second{3.23} & 0.92 & 3.33 & 95.3 & 2.25 \\
\midrule
SAGE w/o VLM & 12.85 & 34.63 & 0.109 & 3.26 & \second{0.86} & 2.73 & 95.9 & 1.47 \\
SAGE w/o Z0 & \textbf{11.74} & \second{30.34} & 0.108 & \textbf{3.08} & \textbf{0.77} & \second{2.11} & \second{96.6} & 0.85 \\
\rowcolor{sagepale}SAGE (ours) & \second{11.79} & 30.49 & 0.108 & \textbf{3.08} & \textbf{0.77} & \textbf{1.73} & \textbf{96.9} & \second{0.69} \\
\bottomrule
\end{tabular}
\begin{tablenotes}[para,flushleft]\footnotesize
\item[] $^\dagger$The prompt contains ROADWork's human scene description.
\end{tablenotes}
\end{threeparttable}

\end{table*}

Fig.~\ref{fig:horizon}(a,~b) shows that the accuracy gain also lies where the control redirects the path. Over the first waypoints, the SAGE variants and constant-velocity extrapolation have nearly the same error, because the observed motion determines the near part of the path, whereas the last five waypoints account for 86\% of the ADE reduction. As the bars in Fig.~\ref{fig:horizon}(b) show, in more than 90\% of turning episodes the demonstrated path departs from its recent motion only in the far third of the horizon, where the candidates diverge and the choice among them determines the error. Consistently, the row-matched lateral error, which covers roughly the nearest 30\,m, differs little among OneVL, the two EMMA variants, and the SAGE variants, from 0.104 to 0.114\,m. Near-field measures therefore understate the differences between methods, which emerge mainly beyond the observed motion.

The SAGE-VLM trajectory yields a more accurate and safer path as an opinion than when it is executed directly: within SAGE, the final displacement is 30.49\,px against 35.86\,px for the trajectory itself, and the hard-violation rate roughly halves. Against OneVL executed directly, SAGE is comparable in accuracy, with an ADE of 11.79 against 12.74\,px and an FDE of 30.49 against 29.02\,px, and it has fewer hard violations, 1.73\% against 2.83\%. On the same backbone, its VLM also completes its output about one-fifth sooner, as Section~\ref{sec:efficiency} reports. SAGE also outperforms each of the four ROADWork-adapted driving VLMs in ADE, FDE, safety score, and hard violations, with about one-third of their hard violations or fewer. This holds even for LightEMMA and OpenEMMA, whose prompts contain the human scene description, so knowing the scene is not sufficient; the limitation lies in converting that knowledge into a precise path, which SAGE delegates to the geometric candidates. The zero-shot Qwen2.5-VL-7B attains the lowest hazard-side share together with the largest ADE among the VLMs, which is why Section~\ref{sec:metrics_consistency} reads these measures together with displacement.

\subsection{Where the gain comes from}
\label{sec:sources}
\label{sec:generation_results}
\begin{figure}[!t]
\centering
\includegraphics[width=0.85\linewidth]{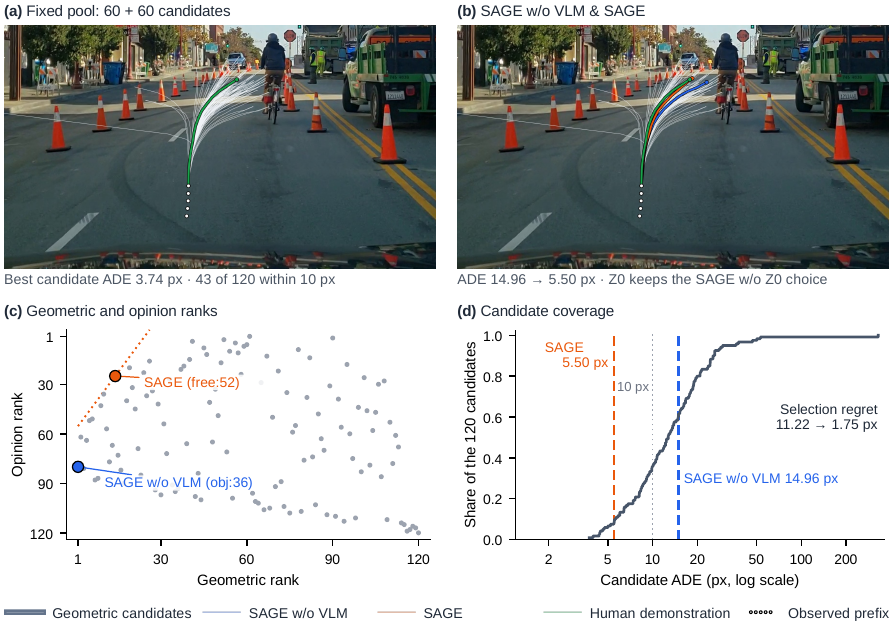}
\caption{Candidate coverage and selection. (a) The 60 + 60 candidates and the demonstration. (b) Choices of SAGE w/o VLM and SAGE. (c) Geometric and opinion ranks; the dotted line marks the fused score of the SAGE w/o Z0 choice, which Z0 retains. (d) Cumulative coverage. Labels such as free:52 give the branch and candidate index.}
\label{fig:candidate_selection}
\end{figure}

\paragraph{Selection is the bottleneck of the geometric layer.}
Table~\ref{tab:ladder} shows that the geometric layer generates close candidates but selects among them weakly. The shared pool contains a path within 10\,px of the demonstration in 87\% of episodes, and its best candidate, at $E^{\mathrm{pool}}=6.04$\,px, is on average about twice as accurate as the 12.85\,px choice of SAGE w/o VLM. The selection regret of Eq.~\eqref{eq:regret} thus accounts for 53\% of the selected path's error. This gap leaves room for a preference that identifies the candidate the control requires. Fig.~\ref{fig:candidate_selection} shows one such case: the pool holds many close candidates, geometry selects a path 14.96\,px away, and the trajectory opinion moves the choice to one 5.50\,px away without changing the pool.

\paragraph{Generation strength and selection strength.}
\label{sec:tao_ablation}
Table~\ref{tab:ladder} also shows that the SAGE-VLM opinion improves every combination of generator and selector, with every interval excluding zero. We pair two generators, the free-space branch alone and the device-aware generator with both branches, with two selectors, the generator's own score and the learned ranker of SAGE w/o VLM. The gain is largest, 1.33\,px, when the pool contains device-guided alternatives and the selection is weak; the learned ranker absorbs part of it, leaving 1.11\,px for SAGE w/o VLM. With the free-space generator, the gain is smaller, which can be attributed to the absence of device-guided alternatives in its pool. On the device-aware pool, the learned ranker, which already uses device evidence, absorbs only 0.22\,px of the opinion's gain, so the trajectory carries information that the device counts and distances do not.

\begin{table}[pos=!t]
\centering
\caption{Generation strength $\times$ selection strength. Bold intervals exclude zero; the shaded row is SAGE w/o VLM.}
\label{tab:ladder}
\footnotesize
\setlength{\tabcolsep}{3pt}
\begin{tabular}{@{}cccccll@{}}
\toprule
\makecell{Device\\roles} & \makecell{Learned\\ranker} & \makecell{$E^{\mathrm{pool}}$\\(px)\,$\downarrow$} & \makecell{Pool within\\10\,px (\%)\,$\uparrow$} & \makecell{ADE (px)\,$\downarrow$\\alone} & \makecell{$\Delta$ADE (px)\,$\downarrow$\\with TAO} & \makecell{$\Delta$ADE (px)\,$\downarrow$\\with anchors} \\
\midrule
\xmark & \xmark & 6.63 & 85.5 & 14.06 & \textbf{$-$0.73 \ci{$-$1.45, $-$0.06}} & $-$0.41 \ci{$-$1.18, $+$0.37} \\
\xmark & \cmark & 6.63 & 85.5 & 12.79 & \textbf{$-$0.88 \ci{$-$1.29, $-$0.52}} & \textbf{$-$0.71 \ci{$-$1.29, $-$0.15}} \\
\cmark & \xmark & 6.04 & 87.4 & 14.58 & \textbf{$-$1.33 \ci{$-$2.20, $-$0.51}} & \textbf{$-$1.24 \ci{$-$2.13, $-$0.40}} \\
\rowcolor{sagepale}\cmark & \cmark & 6.04 & 87.4 & 12.85 & \textbf{$-$1.11 \ci{$-$1.70, $-$0.51}} & $-$0.70 \ci{$-$1.34, $+$0.04} \\
\bottomrule
\end{tabular}

\end{table}

\paragraph{Interface: trajectory or decision anchors.}
The full trajectory is a more informative interface than decision anchors, which keep only the goal and up to two key points. Table~\ref{tab:ladder} shows that its gain is larger than that of anchors on all four generator--selector combinations, by 0.41\,px on SAGE w/o VLM. This advantage can be attributed to the shape that anchors discard, which distinguishes candidates within one maneuver. The gap is largest for SAGE w/o VLM, which combines the device-aware pool with the learned ranker; there, the anchor gain is no longer distinguishable from zero. Once the pool and the selector are both strong, the remaining choice likely lies among paths of one maneuver, which only the shape of the opinion separates.

\paragraph{Results by condition.}

\begin{table}[pos=!t]
\centering
\caption{ADE by episode type and work-zone condition. Bold intervals exclude zero.}
\label{tab:strata}
\footnotesize
\setlength{\tabcolsep}{4pt}
\begin{tabular}{@{}lrrrl@{}}
\toprule
 & & \multicolumn{2}{c}{SAGE, ADE (px)\,$\downarrow$} & \\
\cmidrule(lr){3-4}
Episodes & $n$ & w/o VLM & w/o Z0 & $\Delta$ADE (px)\,$\downarrow$ \\
\midrule
All & 2{,}306 & 12.85 & 11.74 & \textbf{$-$1.11 \ci{$-$1.70, $-$0.51}} \\
\midrule
Laterally departing & 470 & 37.05 & 33.24 & \textbf{$-$3.81 \ci{$-$6.57, $-$0.94}} \\
Not laterally departing & 1{,}836 & 6.66 & 6.24 & \textbf{$-$0.42 \ci{$-$0.54, $-$0.29}} \\
\midrule
Activity area & 1{,}298 & 12.43 & 11.33 & \textbf{$-$1.10 \ci{$-$1.59, $-$0.67}} \\
Transition area & 237 & 9.81 & 8.40 & \textbf{$-$1.41 \ci{$-$3.47, $-$0.29}} \\
Channelized corridor & 457 & 12.76 & 11.38 & \textbf{$-$1.38 \ci{$-$2.75, $-$0.41}} \\
Minimal control & 314 & 17.05 & 16.53 & $-$0.52 \ci{$-$3.29, $+$2.86} \\
\bottomrule
\end{tabular}

\end{table}

Table~\ref{tab:strata} shows that the gain is largest where the temporary control redirects the path: 3.81\,px in laterally departing episodes against 0.42\,px in the others, and 1.10--1.41\,px wherever devices or work activity are visible. Laterally departing episodes make up one-fifth of the evaluation set but account for about 70\% of the ADE reduction. Where the scene shows little temporary control, the gain is small and its interval includes zero, which is consistent with the opinion contributing an interpretation of the control rather than a generic correction of the ranker. The conditions follow a fixed rule on the detections: an activity area if a worker, a police officer, or a work or police vehicle is detected, otherwise a transition area if a taper is detected, that is, at least four channelizing devices on one side that align along a line inclined to the observed heading, otherwise a channelized corridor if channelizing devices are detected, and minimal control otherwise.

\subsection{Reading and bounded intervention}
\label{sec:z0results}
\label{sec:interp_results}
As Table~\ref{tab:reading} shows, MUTCD-structured supervision turns OneVL's trajectory-only network into a reliable reader. When SAGE-VLM names a hazard side, the side agrees with the human description in 92.7\% of cases, and it names the opposite side in only 5.2\% of the 1{,}474 described-side episodes; otherwise, it answers ``none or both'', for which the side predicate abstains. Abstentions therefore outnumber reversals, and the reading errs toward caution: an abstention only forgoes one check, whereas a reversed side would flag a path that respects the control.

\begin{table}[pos=!t]
\centering
\caption{Reading quality. Passage direction is compared with the demonstration on the 339-episode development subset; hazard side uses the 1{,}474 validation episodes whose description names a side.}
\label{tab:reading}
\footnotesize
\begin{tabular}{@{}lccc@{}}
\toprule
 & \makecell{Passage direction\\(\%)\,$\uparrow$} & \makecell{Hazard side agrees\\when named (\%)\,$\uparrow$} & \makecell{Opposite side\\named (\%)\,$\downarrow$} \\
\midrule
SAGE-VLM & 91.2 & 92.7 & 5.2 \\
\bottomrule
\end{tabular}
\end{table}

Table~\ref{tab:z0_effect} shows that the bounded intervention makes the selections more consistent with the control. It lowers the share of choices flagged by $\Gamma$ from 13.1\% to 9.1\%, and on the 177 human-filtered flagged choices, where it acts, it lowers the hard-violation rate from 16.9\% to 12.4\%. It changes 93 of the 2{,}306 selections, all within the displacement bound. Z0 thus resolves about 30\% of the flagged choices and leaves ADE nearly unchanged, 11.79 against 11.74\,px. The remaining flagged choices have no admissible candidate within the score and displacement bounds or lack a usable VLM output, so the bounds limit the reach of the intervention and keep each change local.

\begin{table}[pos=!t]
\centering
\caption{Effect of the bounded intervention Z0. Flagged choices: the 177 human-filtered episodes whose fused choice violates a predicate.}
\label{tab:z0_effect}
\footnotesize
\begin{tabular}{@{}lcc@{}}
\toprule
 & SAGE w/o Z0 & SAGE \\
\midrule
\multicolumn{3}{@{}l}{\emph{All episodes}} \\
\quad Choices flagged by $\Gamma$ (\%)\,$\downarrow$ & 13.1 & 9.1 \\
\quad RSI $>0$ (\%)\,$\downarrow$ & 8.2 & 5.6 \\
\quad CBC $>0$ (\%)\,$\downarrow$ & 2.0 & 1.5 \\
\quad Changed selections & -- & 93 \\
\multicolumn{3}{@{}l}{\emph{Flagged choices}} \\
\quad Hard violations (\%)\,$\downarrow$ & 16.9 & 12.4 \\
\quad Safety score\,$\uparrow$ & 81.1 & 85.2 \\
\bottomrule
\end{tabular}
\end{table}

\paragraph{Human assessment of the changed paths.}
An anonymized human review, summarized in Table~\ref{tab:human_spatial_audit}, confirms these changes in the image. Over the 88 changed cases with definite judgments, visible contact with a worker or device falls from 25 to 13 judgments, and corridor inconsistency from 24 to 13. The changes improve three to four times as many judgments as they worsen, so where the image-space proxies trigger a change, a human reviewer far more often confirms than contradicts it.

\begin{table}[pos=!t]
\centering
\caption{Human review of the paths changed by Z0.}
\label{tab:human_spatial_audit}
\footnotesize
\setlength{\tabcolsep}{3pt}
\begin{tabular}{@{}lrrrrl@{}}
\toprule
Judgment & Before & After & Improved\,$\uparrow$ & Worsened\,$\downarrow$ & Paired change (pp)\,$\downarrow$ \\
\midrule
Corridor inconsistent & 24/88 & 13/88 & 16 & 5 & $-$12.50 \ci{$-$21.11, $-$4.40} \\
Visible contact & 25/88 & 13/88 & 16 & 4 & $-$13.64 \ci{$-$21.79, $-$6.02} \\
\bottomrule
\end{tabular}

\end{table}

\paragraph{OneVL's decoded reasoning as a reader.}
Decoding OneVL's latent reasoning with its auxiliary language decoder is a much slower route to a reading. Table~\ref{tab:latency} shows that it takes about 30\,s per frame, against 0.71\,s for SAGE-VLM's reading and decision.

\subsection{Efficiency}
\label{sec:efficiency}
\label{sec:rq3}

\begin{table}[pos=!t]
\centering
\caption{Per-frame latency at batch size 1 on one idle RTX 6000 Ada GPU.}
\label{tab:latency}
\footnotesize
\begin{tabular}{@{}lrr@{}}
\toprule
Component & Median (ms)\,$\downarrow$ & p90 (ms)\,$\downarrow$ \\
\midrule
\multicolumn{3}{@{}l}{\emph{Geometric layer}} \\
\quad Role segmentation & 2.8 & 2.8 \\
\quad RD-YNet branch, 60 candidates & 258--260 & 260--261 \\
\multicolumn{3}{@{}l}{\emph{SAGE-VLM}} \\
\quad Vision encoder & 34 & 35 \\
\quad Reading and decision & 710 & 867 \\
\quad Goal & 927 & 1{,}077 \\
\quad Complete output & 3{,}083 & 3{,}233 \\
\multicolumn{3}{@{}l}{\emph{Same network and code, back to back}} \\
\quad OneVL trajectory & 3{,}752 & 3{,}781 \\
\quad SAGE-VLM complete output & 3{,}058 & 3{,}189 \\
\multicolumn{3}{@{}l}{\emph{OneVL, released inference code}} \\
\quad Trajectory & 30{,}500 & 30{,}748 \\
\quad Reading via the latent decoder & 30{,}273 & 30{,}370 \\
\bottomrule
\end{tabular}

\end{table}

\begin{figure*}[!t]
\centering
\includegraphics[width=0.9\textwidth]{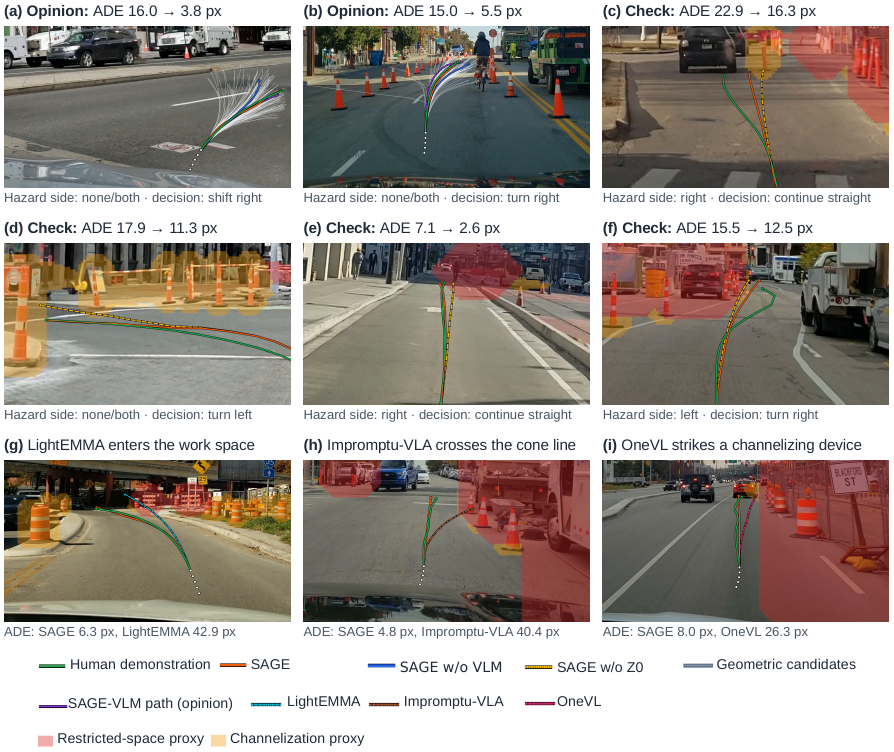}
\caption{Qualitative validation cases. (a, b) The SAGE-VLM opinion selects a close candidate that the geometric ranker missed. (c--f) Z0 replaces a SAGE w/o Z0 choice that enters a shaded restricted-space or channelization proxy with a nearby candidate that does not. (g--i) LightEMMA, Impromptu-VLA, and OneVL violate a hard constraint; SAGE does not. Titles of (a)--(f) give the ADE before and after.}
\label{fig:qualitative}
\end{figure*}

Table~\ref{tab:latency} also shows that the geometric layer never waits for the VLM. Each RD-YNet branch generates its 60 candidates in about 0.26\,s, so a geometric choice is available well before the VLM output, and SAGE-VLM states its reading and decision after 0.71\,s, before its trajectory. On the same Qwen3-VL-4B network and code, SAGE-VLM completes its whole output in 3.06\,s, against 3.75\,s for OneVL's trajectory alone. OneVL's released inference code takes about 30\,s per frame, because its checkpoint configuration disables the key-value cache. The VLM's latency therefore determines how current its opinion is, not whether a path is available.

\begin{figure}[!t]
\centering
\includegraphics[width=0.85\linewidth]{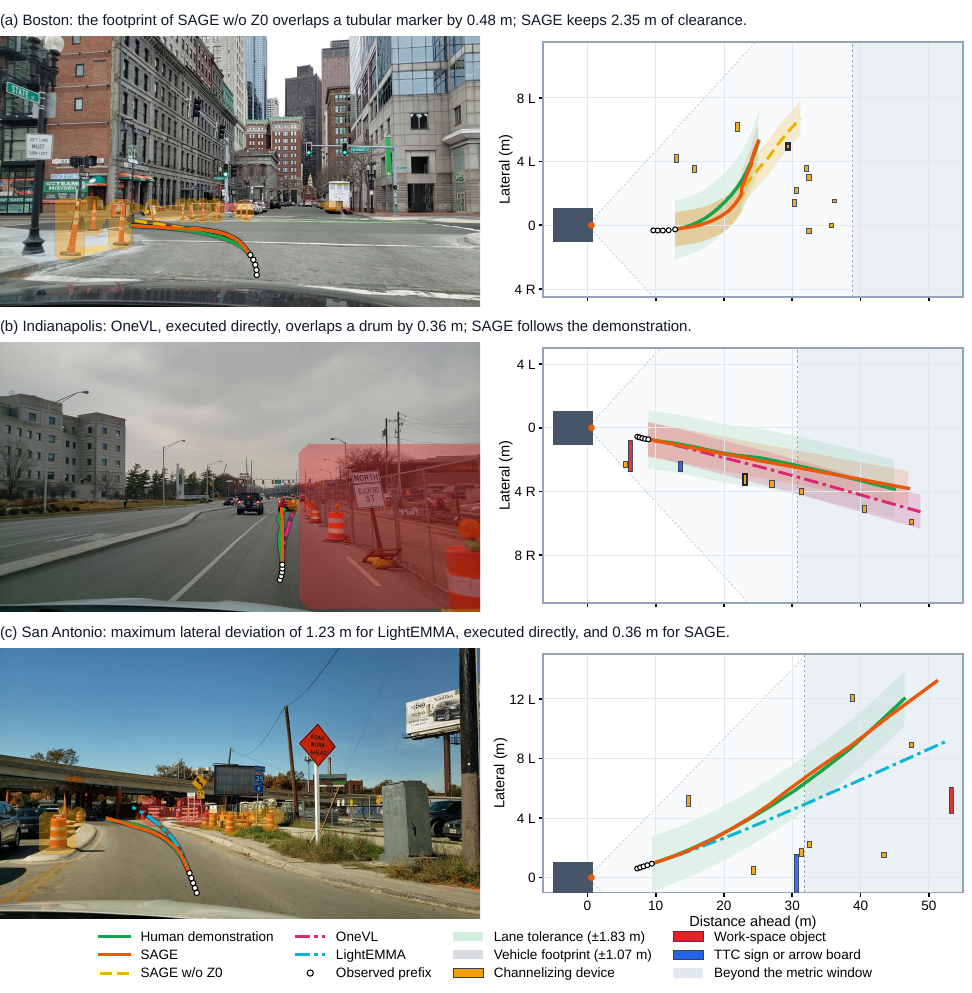}
\caption{Cases (d), (i), and (g) of Fig.~\ref{fig:qualitative} in the device-anchored ground frame. Left: front view. Right: top view with the $\pm1.83$-m lane tolerance around the demonstration and the $\pm1.07$-m vehicle footprint. The lateral axis is stretched about twofold; shading marks distances beyond the evaluated metric-frame rows.}
\label{fig:bev}
\end{figure}

\subsection{Qualitative analysis}
\label{sec:qualitative}

\begin{figure}[!t]
\centering
\includegraphics[width=0.85\linewidth,height=0.78\textheight,keepaspectratio]{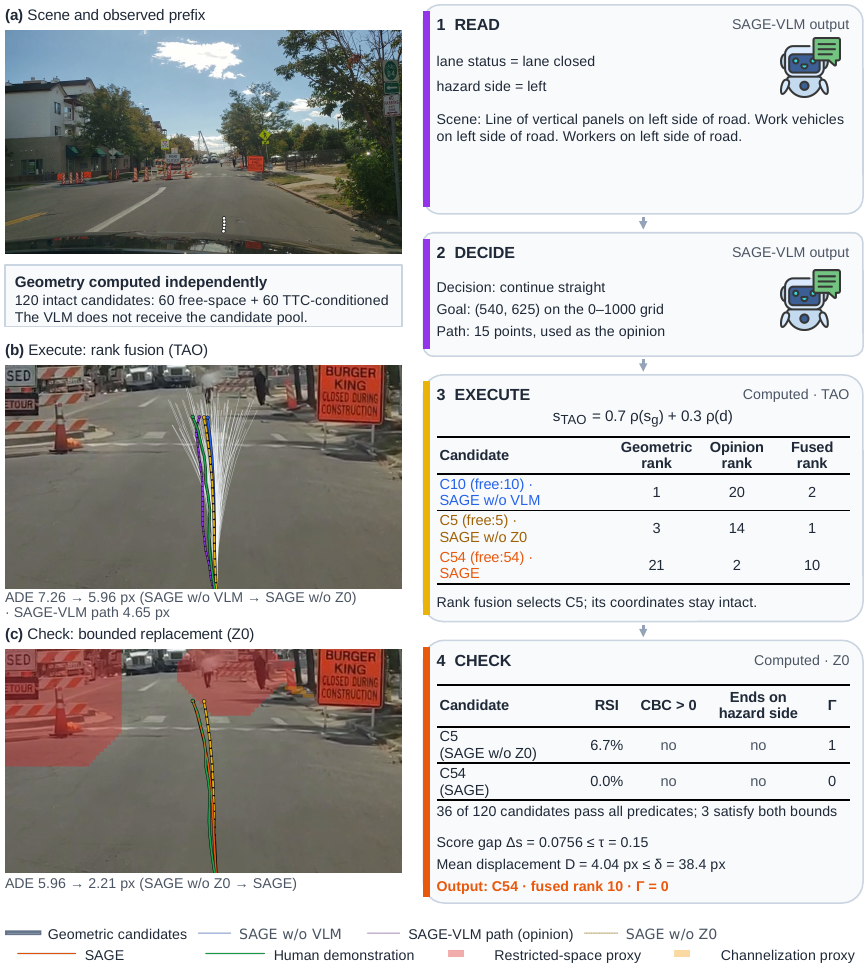}
\caption{Four-stage trace for one Denver frame: (a) scene and observed prefix, (b) TAO rank fusion, (c) Z0 bounded replacement. Read and Decide show recorded VLM output; Execute and Check show computed ranks and predicates. TAO changes the SAGE w/o VLM choice C10 to C5, and Z0 replaces C5 with C54. Ranks are one-based; candidate indices are zero-based.}
\label{fig:reasoning_trace}
\end{figure}

Fig.~\ref{fig:qualitative} illustrates the two mechanisms; green marks the human demonstration, blue the choice of SAGE w/o VLM, dashed yellow that of SAGE w/o Z0, orange the SAGE output, and dashed purple the SAGE-VLM path. In panels (a) and (b), the pool contains a candidate close to the demonstration, but the geometric ranker selects one that diverges in the far field; SAGE-VLM decides to shift or turn right, and its trajectory moves the selection to the close candidate. In panels (c)--(f), the choice of SAGE w/o Z0 enters the red restricted-space proxy or the amber channelization proxy, and Z0 replaces it with a nearby candidate that stays clear, which in all four cases also lies closer to the demonstration. In panels (g)--(i), LightEMMA, Impromptu-VLA, and OneVL, executed directly, enter the work space, cross the cone line, or strike a channelizing device, whereas SAGE stays in the traffic space.

Fig.~\ref{fig:bev} shows three of these cases in the device-anchored ground frame, where the differences become clearance and lane-keeping margins. In panel (a), the footprint of SAGE w/o Z0 overlaps a tubular marker by 0.48\,m, whereas SAGE keeps 2.35\,m of clearance; in panel (b), the footprint of OneVL overlaps a drum by 0.36\,m; and in panel (c), LightEMMA deviates from the demonstration by up to 1.23\,m laterally, against 0.36\,m for SAGE. These overlaps, 0.36 to 0.48\,m, are small against the 1.83-m half-lane tolerance, so a path can stay within lane-level accuracy and still strike a device; this is why the evaluation reports collisions in addition to lateral error.

Fig.~\ref{fig:reasoning_trace} traces one Denver frame through the four stages, separating the emitted VLM fields from the computed ranks and predicates. The SAGE output is a geometric candidate that the opinion ranks second, and it lies closer to the demonstration than the SAGE-VLM path itself, at 2.21 against 4.65\,px. The opinion therefore need not be the most accurate path: it only needs to rank a close candidate highly, and the geometric pool supplies the precise coordinates.

\section{Discussion}
\label{sec:discussion}

\subsection{Implications for automated driving in work zones}
The central transportation problem in work zones lies in converting a temporary-control arrangement into a particular passing path. SAGE addresses this conversion through a division of labor in which each component handles the part of the problem it represents best. Functional roles enter the generator, detector and path features enter the geometric ranker, the VLM's trajectory enters as a preference over candidates, and the VLM's reading enters a standard-grounded check with bounded authority. Executed directly, the SAGE-VLM trajectory ends farther from the demonstration than when it enters as a rank opinion, which also lowers the hazard-side share, and its reading, entered into the bounded intervention, reduces the choices that the instantiated provisions flag (Sections~\ref{sec:rq2} and~\ref{sec:z0results}). Where a foundation-model output enters the system can thus matter as much as which model produces it. Because the geometric pool already contains close alternatives, the remaining difficulty is selection, and each output enters where it can inform that selection without constructing coordinates.

The results also indicate where a VLM adds most to a planner. The same opinion helps most when the planner generates good alternatives but selects among them weakly, and a learned ranker absorbs part of what the VLM contributes, while a clear gain remains on top of it. For developers, this suggests measuring selection regret, i.e., the gap between the selected candidate and the best one in the pool, before integrating a VLM: a VLM trajectory opinion pays off in a planner whose pool already contains the required path but whose selection misses it. The gain also concentrates in the far part of the horizon, where the devices redirect the path and the candidates diverge, so evaluations of work-zone planners benefit from reporting far-field errors separately. For a well-tuned planner, the case for a VLM also rests on what the planner cannot provide, namely a stated reading of the control that a standards check can use.

A third implication concerns how such a VLM is obtained. Driving VLAs trained to a single trajectory format can lose the ability to follow other instructions, as the released OneVL checkpoint shows. A reading then cannot be requested by prompting, and reading it out of latent reasoning takes tens of seconds per frame. MUTCD-structured supervision restores the reading with one fine-tuning run and keeps the ADE of the driving checkpoint's trajectory (12.15 against 12.74\,px), which suggests that targets structured by the governing standard offer a practical way to make a VLM state the control on which its decision rests.

\subsection{Implications for traffic control practice and oversight}
The results also inform how temporary traffic control can accommodate AVs. MUTCD Part~5 identifies the intolerance of driving automation systems to non-uniform device design and application as a limiting factor, and for TTC zones it recommends consistent sign placement, wider retroreflective material on, or reduced spacing of, channelizing devices, and continuous markings in lane transitions \citep[\S5A.04, \S5B.04]{fhwa2025mutcd}. Our stratified results locate where interpretation matters most for current models: the opinion gains most where the path departs laterally from its heading (3.81\,px), and it gains 1.10--1.41\,px wherever channelizing devices or work activity are visible. These layouts are natural candidates for evaluating such uniformity measures with AV perception in the loop, and the device-anchored frame allows the outcome to be reported in lane widths and lateral offsets.

Infrastructure data can complement on-board interpretation. A WZDx feed describes a work zone through lane-status values such as closed, shift-left, and merge-right \citep{wzdx2024spec}, and the reading of SAGE-VLM uses a comparable vocabulary of lane statuses and hazard sides, and its decisions name shifts. Accordingly, a feed, when available and current, could enter the reading as additional evidence or supervise it. Roadside units can also measure the layout itself: ultra-wideband anchors mounted on cones reconstruct the work-zone geometry from vehicle-to-infrastructure ranges \citep{liu2026v2iuwb}, which could supply the channelization boundary that the checks now take from the image. Conversely, a disagreement between the feed and the on-board interpretation could flag a layout that changed after publication, a situation that the MUTCD anticipates because TTC is to be modified as work progresses \citep[\S6A.02]{fhwa2025mutcd}.

Finally, auditable decisions support the oversight of AVs in work zones. The MUTCD asks agencies to monitor crash records in TTC zones to identify needed changes \citep[\S6A.02]{fhwa2025mutcd}, and the recall discussed in Section~\ref{sec:intro} shows a developer restricting operation after its system misread temporary control \citep{nhtsa2026recall}. For each frame, SAGE records the VLM's reading and decision, the predicates evaluated on the selected path, and the decision source. Such a record gives agencies and developers a concrete account of why a vehicle chose its path through a TTC zone and can help diagnose failures such as those reported in that recall.

\subsection{Limitations and future research}
\label{sec:limitations}
Several research directions follow from this work. First, the present evaluation compares predicted paths with real-world human demonstrations in open loop. Closed-loop evaluation in work-zone scenarios, for instance in simulations reconstructed from real dashcam logs \citep{ghosh2026dash2sim}, would show how the selected paths interact with other road users over time. Second, the instantiated provisions are those that the image-space pathway representation can express. Metric and time-resolved ground truth would allow the framework to check provisions that depend on depth and speed, such as buffer and taper lengths and flagger control, and to report time-based surrogate safety measures such as time to collision and post-encroachment time. Third, VLMs that emit their reading and path within the time that a passage decision stays valid, for instance through shorter outputs or earlier emission of the path, would let the opinion follow rapidly changing scenes. Fourth, agency work-zone feeds such as WZDx could enter the reading as additional evidence, and extending the framework to the standards of other jurisdictions and to high-speed, rural, and nighttime work zones would broaden its coverage; nighttime crashes account for more than half of the fatal work-zone crashes of 2024 \citep{fhwa2026nwzaw}.

\section{Conclusion}
\label{sec:conclusion}
In work zones, channelizing devices, signs, and arrow boards temporarily override the permanent lane geometry, and an AV must follow the path they designate within a fraction of a lane width while keeping clear of the work space. This paper presented SAGE, a read--decide--execute--check framework that connects a vision--language model's reading of the temporary control to precise work-zone pathway prediction. The key idea of SAGE lies in decoupling path construction, path preference, and control compliance around a shared candidate space. Specifically, a role-decoupled geometric generator proposes intact candidate pathways, and a geometric ranker orders them by default. SAGE-VLM, trained with MUTCD-structured supervision, reads the lane status and the hazard side, states its passage decision, and contributes its trajectory as an opinion over the candidates, subject to a training-split admission test. MUTCD temporary-traffic-control provisions, instantiated as checkable predicates, drive a bounded intervention that uses the VLM's reading. Every output remains one intact geometric candidate.

Experiments on 2{,}306 real-world ROADWork validation episodes demonstrate the effectiveness of the framework. The geometric candidates contain a path within 10\,px of the human demonstration in 87\% of episodes, yet selection regret accounts for 53\% of the selected path's error, so candidate selection is the bottleneck. SAGE reduces ADE and FDE by 8.3\% and 12.0\% relative to the geometric ranker, reduces the share of paths that violate the evaluated hard constraints by 37\%, and halves the share of paths that end on the described hazard side in episodes whose demonstration ends elsewhere. The opinion consistently improves every combination of generator and selector that we tested, most when the pool contains device-guided alternatives and selection is weak. The gain concentrates in the far part of the horizon, where the temporary control redirects the path. Notably, SAGE achieves accuracy comparable to the released ROADWork checkpoint of OneVL, its VLM completes its output about one-fifth sooner with the same Qwen3-VL-4B backbone, and it outperforms four other adapted driving VLMs in displacement error and safety measures.

These findings suggest a practical division of labor for work-zone automation: geometry constructs the action space, a VLM that has been taught to read the control refines the choice within it, and codified standards bound any intervention. For road agencies, the same structure ties each automated decision to a stated reading and to the MUTCD provisions evaluated on it, using a vocabulary of lane statuses and hazard sides comparable to that of agency work-zone feeds. While the present evaluation uses real-world driving data in open loop, future work will evaluate the framework in closed loop, train VLMs that read and predict accurately within the time a decision stays valid, and extend the evaluation to additional work-zone datasets.

\FloatBarrier
\section*{CRediT authorship contribution statement}
\textbf{Jiaxi Liu:} Conceptualization, Data curation, Formal analysis, Investigation, Methodology, Project administration, Resources, Software, Validation, Visualization, Writing -- original draft, Writing -- review \& editing. \textbf{Keshu Wu:} Supervision, Writing -- review \& editing. \textbf{Weizhe Tang:} Writing -- review \& editing. \textbf{Bofeng Cao:} Writing -- review \& editing. \textbf{Sikai Chen:} Supervision, Writing -- review \& editing. \textbf{Xiaopeng Li:} Supervision, Writing -- review \& editing. \textbf{Steven T. Parker:} Funding acquisition, Supervision, Writing -- review \& editing. \textbf{Bin Ran:} Funding acquisition, Supervision, Writing -- review \& editing.
\section*{Acknowledgments}
This study is supported by the Center for Connected and Automated Transportation (CCAT), University of Michigan, through Grant No. 69A3552348305. The views expressed are those of the authors.
\section*{Declaration of competing interest}
The authors declare that they have no known competing financial interests or personal relationships that could have appeared to influence the work reported in this paper.
\section*{Data availability}
ROADWork is available from its original release \citep{ghosh2025roadwork}. The candidate records, stored model outputs, SAGE-VLM adapter, evaluation scripts, and checkpoint-identification manifest are intended for release with the final publication. The project page, with walkthrough videos and interactive views of the evaluated cases, is available at \url{https://hhjj233.github.io/SAGE-page/}.
\section*{Declaration of generative AI and AI-assisted technologies in the manuscript preparation process}
During the preparation of this work the authors used ChatGPT (OpenAI) and Claude (Anthropic) to edit language and to assist with editing text and writing code for data processing and figures. After using these tools, the authors reviewed and edited the content as needed and take full responsibility for the content of the publication.
\FloatBarrier
\bibliographystyle{cas-model2-names}
\bibliography{refs}
\end{document}